\documentclass{article}

 \usepackage[sglblindworkshop, final]{neurips_2026}
\workshoptitle{Lightweight, Trustworthy Foundation Models (LIGHT)}

\usepackage{hyperref}
\usepackage{url}
\usepackage{booktabs}       
\usepackage{amsfonts}       
\usepackage{nicefrac}       
\usepackage{microtype}      
\usepackage{adjustbox}
\usepackage{float}
\usepackage{wrapfig}
\usepackage{amsmath}
\usepackage{amssymb}
\usepackage{multirow}
\usepackage[table]{xcolor}
\usepackage{colortbl}
\definecolor{lightmaroon}{RGB}{242, 222, 222}
\definecolor{darkgreen}{RGB}{0, 100, 0}
\title{Explainability-Guided History Distillation for Efficient Sequential Recommendation}

\author{%
  Akash Pandey\thanks{Corresponding author.} \\
  Capital One, AI Foundations \\
  San Jose, CA, USA \\
  \texttt{akash.pandey@capitalone.com} \\
  \And
  Addrish Roy \\
  Capital One, AI Foundations \\
  New York, NY, USA \\
  \texttt{addrish.roy@capitalone.com} \\
  \And
  Kanisha Shah \\
  Capital One, AI Foundations \\
  New York, NY, USA \\
  \texttt{kanishaarpit.shah@capitalone.com} \\
  \And
  Akshat Shreemali \\
  Capital One, AI Foundations \\
  New York, NY, USA \\
  \texttt{akshat.shreemali@capitalone.com} \\
    \And
  Dwipam Katariya \\
  Capital One, AI Foundations \\
  McLean, VA, USA \\
  \texttt{dwipam.katariya@capitalone.com} \\
  \And
  Swapnil Shinde \\
  Capital One, AI Foundations \\
  McLean, VA, USA \\
  \texttt{swapnil.shinde@capitalone.com} \\
  \And
  Amanda Ding \\
  Capital One, AI Foundations \\
  San Jose, CA, USA \\
  \texttt{amanda.ding@capitalone.com} \\
  \And
  Kalanand Mishra \\
  Capital One, AI Foundations \\
  San Jose, CA, USA \\
  \texttt{kalanand.mishra@capitalone.com} \\
  \And
  Pranab Mohanty \\
  Capital One, AI Foundations \\
  Seattle, WA, USA \\
  \texttt{pranab.mohanty@capitalone.com} \\
}

\begin{document}

\maketitle

\begin{abstract}
Sequential recommendation models, spanning self-attention based Transformer architectures to state-space models, rely on processing complete interaction histories, incurring inference costs that scale with sequence length. While efficient attention variants and heuristic truncation mitigate this computational footprint, they overlook internal model attribution signals and neglect the efficiency gains achievable by distilling the raw interaction sequence itself. To bridge this gap, we propose explainability-driven sequence distillation, a retraining-free framework that leverages post-hoc attribution methods to score each interaction's predictive contribution. By retaining only the highest-scoring positions, our framework compresses user histories offline, enabling shorter sequences to accelerate online inference without altering underlying model parameters. Across four dataset configurations and four architectural paradigms, including autoregressive, bidirectional, state-space, and long-sequence backbones, Integrated Gradient-based sequence distillation preserves or improves ranking quality while substantially reducing inference GFLOPs, outperforming recency-based truncation by an average of 11.5\% across all evaluated settings. Crucially, even on benchmarks with extreme temporal recency bias, our approach matches or exceeds recency truncation in the performance-efficiency trade-off, an advantage that widens under noisy input sequences. These findings demonstrate that post-hoc explainability, traditionally restricted to model interpretation, serves as a general, retraining-free principle for distilling raw input contexts and reducing online serving costs in sequential recommendation.
\end{abstract}

\section{Introduction}
Sequential recommendation models predict which item a user will interact with next, given the chronological sequence of their past interactions. This
  formulation is central to modern recommender systems: e-commerce platforms, streaming services, and short-video applications all rely on a user's recent
  behavior to rank candidate items for their next session. The input to these models is a sequence of item identifiers, ordered by time, where each identifier
  corresponds to a past user interaction, such as a purchase, a view, or a click, truncated or padded to a fixed context window of length $L$. Each item is
  mapped to a learned embedding vector, and the model processes the resulting sequence of embeddings to produce a score over the item catalog. Modern architectures such as
  SASRec~\citep{kang2018sasrec} and BERT4Rec~\citep{sun2019bert4rec} implement this mapping using self-attention, allowing every position in the sequence to
  directly attend to every other position, which has proven highly effective at capturing long-range dependencies in user behavior.

  This effectiveness comes at a cost. Self-attention scales quadratically with sequence length $L$: both memory and inference compute grow as
  $\mathcal{O}(L^2)$. As platforms retain longer interaction histories, whether because users have been active for years or because a domain naturally
  produces high-frequency interactions ~\citep{fintrec}, this quadratic scaling becomes a deployment bottleneck. Serving predictions at low latency and low cost is a
  practical requirement, and naively feeding an ever-growing history into a self-attention model is increasingly incompatible with
  that requirement.

  A natural way to reduce this cost is to shorten the sequence before inference ~\citep{fintrec}. Efficient attention variants, such as linear~\citep{linear_attn} and  sparse~\citep{child2019generating} attention, reduce the cost of processing a fixed-length sequence, but they do not decide which interactions to keep when  the sequence itself must be truncated. Simple heuristics fill this gap. Recency-based truncation keeps only the most recent $K$ interactions. Random  truncation keeps $K$ interactions at random. Neither heuristic uses any signal from the trained model, and both implicitly assume that recency, or arbitrary  coverage, is a good proxy for relevance. This assumption does not hold universally: prior work shows that sequential recommendation models concentrate  their predictive signal on a small, dataset-dependent subset of interactions~\citep{rao2021dynamicvit, he}, and, as we show later, this subset is not always  the most recent one.

  Storing and caching interaction data for online inference is computationally expensive and memory-intensive, particularly as sequence lengths scale. This raises a natural question: can we identify the informative subset of interactions using signals from the model itself, without retraining it? Post-hoc explainability methods are designed exactly for this purpose. Given a trained model and an input, they assign a score to each input position reflecting its contribution to the prediction. By computing these scores offline to prune interaction histories before deployment, we pay the explanation cost once in storage. Online inference then operates on significantly shorter sequences, simultaneously drastically reducing inference GFLOPs.

\noindent\textbf{Our Contributions.} We introduce explainability-guided sequence distillation, a \textbf{model-agnostic}, \textbf{retraining-free framework} that integrates post-hoc attribution into the inference pipeline of sequential recommendation models. Rather than serving full interaction histories, our framework retains only the highest-scoring positions identified by an explainability method, \textbf{compressing sequences offline} to drastically \textbf{accelerate online inference} without altering underlying model parameters. Our framework seamlessly supports any post-hoc attribution method that yields position-wise feature importance and generalizes across diverse sequential recommendation architectures. Our key contributions are:
\begin{itemize}
    \item We formalize sequence distillation as an architecture-agnostic position selection problem guided by per-step attribution scores, introducing a normalized AUC metric to quantify performance across the compression spectrum alongside Pareto trade-off analyses at optimal operating points.
    \item We perform an extensive empirical evaluation across four backbone architectures and four dataset configurations, demonstrating that attribution-guided distillation consistently outperforms standard recency truncation and random pruning, achieving superior performance-efficiency trade-offs.
    \item Through comprehensive ablations, we demonstrate the robustness of explainability-guided distillation under sequence corruption, while characterizing its sensitivity to model capacity and distinct attention mechanisms (standard, linear, and sparse).
\end{itemize}
  
\section{Related Works}
\vspace{-3mm}
\textbf{Architecture-Based Efficient Models.} 
Mitigating the quadratic complexity $\mathcal{O}(L^2)$ of self-attention in sequential recommendation has driven extensive architectural research. Approaches include linearizing attention via kernel operations to achieve $\mathcal{O}(L)$ complexity~\citep{linear_attn}, restricting attention to localized or strided windows via sparse patterns~\citep{child2019generating}, and projecting context dimensions via low-rank approximations~\citep{wang2020linformer}. More recently, selective state space models (SSMs) like Mamba have emerged as linear-time alternatives to Transformers for long user sequences~\citep{gu2023mamba}. However, while these structural modifications reduce the compute footprint per step, they process fixed context windows uniformly and cannot identify or prune uninformative interactions from the input history itself.

\textbf{Patching in Time Series and Sequential Modeling.} 
Patching compresses sequences by grouping contiguous time steps into local sub-sequences. Originally popularized in time series forecasting to preserve temporal structure (e.g., PatchTST~\citep{nie2022time}), patching has recently been adapted to sequential recommendation, particularly with Large Language Model (LLM) backbones such as PatchRec, which aggregates textual item tokens and sessions into multi-grained patches directly at the raw input prompt level during online inference~\citep{patchrec}. However, for complex temporal user interactions comprising a mixture of categorical and numerical features, such patching must typically be performed online in the latent feature space, leaving the domain of raw input-level sequence distillation unexplored.

Crucially, raw sequence-level distillation is orthogonal to architectural efficiency and can be seamlessly combined with the models discussed above; we empirically demonstrate this synergy in our results by evaluating our proposed framework across standard, linear, and sparse attention mechanisms.

\begin{figure}[H]
      \centering
      \includegraphics[width=0.9\textwidth]{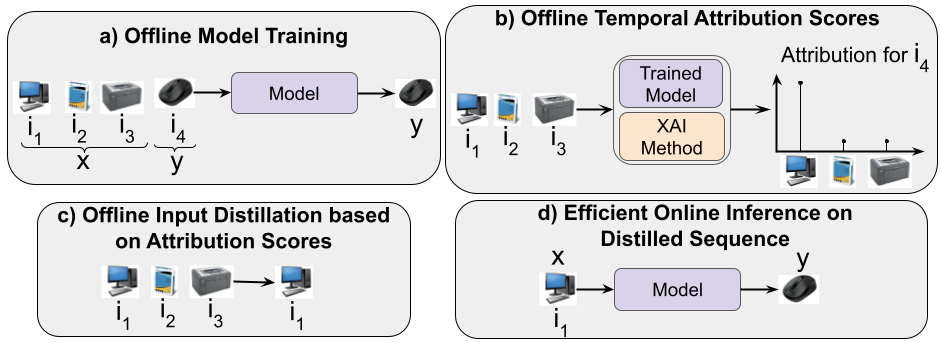}
      \vspace{-4mm}
      \caption{Framework pipeline. (a) Offline training on full history. (b) Post-hoc attribution scoring using a trained model and XAI method. (c) Offline sequence distillation retaining top-attributed items ($i_1$). (d) Efficient online inference on distilled history with the frozen model.}
      \label{fig:main_fig}
  \end{figure}
  
\section{Methodology}\label{method}
\vspace{-3mm}
\subsection{Problem Formulation}
Let $\mathcal{U}$ and $\mathcal{I}$ denote the user and item sets, with $|\mathcal{U}|$ users and $|\mathcal{I}|$ items. For a user $u \in \mathcal{U}$, we
  observe a chronologically ordered interaction sequence $\mathcal{S}_u = (i_1, \dots, i_L)$ truncated to a fixed context window of length $L$. An embedding
  matrix $\mathbf{E} \in \mathbb{R}^{|\mathcal{I}| \times d}$ maps each item to a $d$-dimensional vector $\mathbf{e}_t \in \mathbb{R}^d$, yielding the input
  representation
  \begin{equation}
  \mathbf{X} = \bigl[\mathbf{e}_1;, \mathbf{e}_2;, \dots;, \mathbf{e}_L\bigr] \in \mathbb{R}^{L \times d}.
  \end{equation}
  Depending on the dataset, $\mathbf{E}$ may encode raw item features (e.g., genre tags, interaction counts) or serve as a learnable lookup table trained
  end-to-end with the model. A sequential recommendation model $f_{\theta}: \mathbb{R}^{L \times d} \to \mathbb{R}^{\gamma}$ maps the input sequence to a score
  vector $\hat{\mathbf{y}} \in \mathbb{R}^{\gamma}$. For binary or multi-class prediction tasks (e.g., whether a user will engage with a given video), $\gamma =
  C$ where $C$ is the number of classes; for next-item prediction, $\gamma = |\mathcal{I}|$.

  Not all positions in $\mathcal{S}_u$ contribute equally to the model's decision. Prior work has shown that sequential recommendation models concentrate their
  predictive signal on a small subset of interactions~\citep{rao2021dynamicvit, he}. In practice, interaction histories are stored offline and served at
  inference time; this raises a natural question: \emph{can we distill the stored interaction sequence offline, prior to deployment, using post-hoc explainability
  methods, such that only the most informative time steps are retained for online inference, reducing computational cost without retraining the model?}


To this end, we employ a post-hoc explanation method $g(f_\theta, \mathbf{X}, c) \to \boldsymbol{\phi} \in \mathbb{R}^L$ that assigns a scalar attribution score $\phi_j$ to each position $j \in [1, L]$ with respect to a target class or item $c$. A positive $\phi_j$ indicates that the interaction at time step $j$ increases the predicted probability $\hat{y}_c$, while a negative $\phi_j$ suppresses it. Crucially, this attribution-guided distillation is performed strictly offline; during online serving, newly arriving interaction data is appended directly to the sequence and subsequently distilled in the next offline cycle. We empirically demonstrate that retaining only the high-attribution time steps either preserves predictive performance or, in many cases, improves it by removing noisy and irrelevant interactions, while simultaneously reducing GFLOPs. The complete end-to-end framework, from offline attribution scoring to efficient online inference, is illustrated in Figure~\ref{fig:main_fig}.


\subsection{XAI-driven Distillation of Sequence}\label{method_description}
\vspace{-2mm}
 \noindent\textbf{Step 1: Training on the full sequence.} We train a transformer-based sequential recommendation model $f_\theta$ on the full interaction sequence $\mathbf{X} \in \mathbb{R}^{L \times d}$ with context window length $L$ kept intact. Concretely, we consider SASRec~\citep{kang2018sasrec} and BERT4Rec~\citep{sun2019bert4rec} as representative architectures, both
  of which employ multi-head self-attention over the full sequence. The computational cost of a single forward pass scales as $\mathcal{O}(L^2 d)$ in the
  attention layers, making inference latency sensitive to the sequence length $L$.

\noindent\textbf{Step 2: Computing temporal attribution scores.} Once $f_\theta$ is trained and frozen, we apply a post-hoc explanation method $g(f_\theta, \mathbf{X}, c) \to \boldsymbol{\phi} \in \mathbb{R}^L$, where $\phi_j$ quantifies the contribution of the interaction at position $j \in [1, L]$ to the predicted score $\hat{y}_c$, with $c = \arg\max_{c} \hat{\mathbf{y}}$ being the top predicted item and $\hat{\mathbf{y}} \in \mathbb{R}^{|\mathcal{I}|}$ the predicted score vector over all items.

\noindent\textbf{Step 3: Distilling the sequence.} Given attribution scores $\boldsymbol{\phi} \in \mathbb{R}^L$, we select the indices of the top-$K$ highest attribution scores and sort them chronologically to preserve causal ordering:
\begin{equation}
    \mathcal{S}_K = \operatorname{sort}\left(\operatorname{ArgTopK}(\boldsymbol{\phi}, K)\right), \qquad K \in \{1, \dots, L\}.
\end{equation}
The distilled input sequence $\mathbf{X}_{\mathcal{S}_K} \in \mathbb{R}^{K \times d}$ is then constructed by selecting the corresponding vectors:
\begin{equation}
    \mathbf{X}_{\mathcal{S}_K} = [\mathbf{x}_j]_{j \in \mathcal{S}_K},
\end{equation}
which retains only the $K$ most informative interactions. Note that Steps 2 and 3 are executed offline during user history storage, enabling efficient online inference.

\noindent\textbf{Step 4: Inference on the distilled sequence.} During the online inference stage, the frozen model $f_\theta$ is evaluated directly on $\mathbf{X}_{\mathcal{S}_K}$:
  \begin{equation}
      \hat{\mathbf{y}}_{\mathcal{S}_K} = f_\theta\left(\mathbf{X}_{\mathcal{S}_K}\right) \in \mathbb{R}^{|\mathcal{I}|}.
  \end{equation}
Reducing input sequence length from \(L\) to \(K\) lowers online inference overhead across paradigms: for self-attention backbones (SASRec, BERT4Rec), quadratic layer costs decrease from \(\mathcal{O}(L^2 d)\) to \(\mathcal{O}(K^2 d)\), yielding a compute reduction factor of \((K/L)^2\); for state-space architectures (Mamba), compute decreases linearly by a factor of \(K/L\).

\subsection{Choice of XAI Methods}
\vspace{-2mm}

We consider four post-hoc XAI methods that are compatible with transformer-based sequential recommendation architectures. \textbf{Integrated Gradients
  (IG)}~\citep{ig} attributes predictions by integrating gradients from a zero baseline to the input.
  \textbf{GradientSHAP}~\citep{shap} approximates SHAP values via expected gradients over noise-perturbed baselines. \textbf{Attention
  tracing}~\citep{attentiontracing} propagates attention weights across layers to obtain a single attribution per input position without requiring an
  additional backward pass. \textbf{Grad-SAM} (Attention$\times$Gradient)~\citep{gradsam} combines attention weights with their gradients to highlight positions that are both
  structurally attended to and gradient-relevant. These four methods span gradient-based, Shapley-inspired, and attention-based attribution families, providing a
  diverse evaluation of temporal relevance in sequential recommendation. The implementation details for XAI methods are given in Appendix \ref{app:ig_grad}.

Methods such as LIME~\citep{lime} and DeepLIFT~\citep{deeplift} were excluded: the former is prohibitively slow at long sequence
  lengths, and the latter is incompatible with the residual connections present in transformer architectures.
  
\subsection{Metrics to Quantify the Benefit of Sequence Distillation} \label{metrics}
To evaluate the effect of sequence distillation across all compression levels, we consider different values for $K : K\in[1, L]$ and define two summary metrics based on the area under the
  distillation curve.

 \noindent\textbf{Normalised AUC.} For a given method and metric $m \in \{\text{HR@10}, \text{NDCG@10}\}$, we first normalise the metric at each distillation level $K$
  by its value at the full sequence length $L$:
  \begin{equation}
      \tilde{m}(K) = \frac{m(K)}{m(L)},
 \label{normalized_eq}
  \end{equation}
  so that $\tilde{m}(L) = 1$ by construction. We then compute the area under the curve $\tilde{m}(K)$ over all evaluated $K$ values via the trapezoidal rule,
  normalised by the range of $K$:
  \begin{equation}
      \mathrm{AUC}_{m} = \frac{1}{L - 1} \int_{1}^{L} \tilde{m}(K) \mathrm{d}K.
  \label{eq:auc_metric}
  \end{equation}
  A value of $\mathrm{AUC}_m > 1$ indicates that, at some distilled sequence length, the model performs better than on the full sequence ($K=L$),
    while $\mathrm{AUC}_m < 1$ indicates the performance degraded after distillation. Overall, a higher $\mathrm{AUC}_m$ for one method relative to another
    indicates that it performs better after distillation. This metric captures the overall quality of a distillation method across the entire compression spectrum
    rather than at a single operating point.

\noindent\textbf{Performance--efficiency trade-off.} To assess the practical efficiency gains achievable across distillation strategies without introducing test-set selection bias, we complement the holistic \(\mathrm{AUC}_m\) metric with a two-dimensional trade-off analysis mapping operational ranking quality against inference compute. Specifically, for each method, the target sequence length is selected strictly a priori by sweeping over \(\mathcal{D}_{\mathrm{valid}}\): \(K^{*} = \arg\max_{K < L} m_{\mathcal{D}_{\mathrm{valid}}}(K)\). Note that \(K^{*}\) is determined globally at the run/seed level for each method rather than adaptively per sample, reflecting a fixed-length operational deployment budget. We then report the performance of this fixed operating point on the held-out test set \(\mathcal{D}_{\mathrm{test}}\), plotting absolute test performance against raw inference compute via the coordinate tuple \(\left(\mathrm{GFLOPs}(K^{*}), m_{\mathcal{D}_{\mathrm{test}}}(K^{*})\right)\). This formulation captures a realistic deployment setting where hyperparameter selection is strictly decoupled from test evaluation. Methods situated toward the \textbf{upper-left region} achieve an optimal trade-off, recovering the full-sequence predictive capacity while incurring only a fraction of the serving latency. Crucially, all reported GFLOPs calculations comprehensively account for both the backbone sequential encoding pass and the final projection onto the complete item catalog (the \(\mathcal{O}(|\mathcal{I}| \cdot d)\) scoring layer), ensuring a fair and faithful representation of end-to-end serving cost.
  
\subsection{Datasets}\label{datasets}
\vspace{-2mm}
  We evaluate our approach on three public benchmarks: KuaiRand-1k~\citep{gao2022kuairand}, MovieLens-1M, and MovieLens-10M~\citep{harper2015movielens}. Each dataset is partitioned into $\mathcal{D}_{\text{train}}$, $\mathcal{D}_{\text{valid}}$, and $\mathcal{D}_{\text{test}}$, with all evaluation performed on $\mathcal{D}_{\text{test}}$ under the next-item prediction task (ranking the full item vocabulary given an interaction sequence). KuaiRand-1k is a short-video interaction dataset evaluated at two sequence lengths, $L=500$ and $L=1000$ (KuaiRand-1k-LONG), to analyze explainability behavior over extended interaction horizons. MovieLens-1M and MovieLens-10M are movie rating benchmarks evaluated at $L=200$ and $L=500$, respectively, following standard conventions~\citep{kang2018sasrec, sun2019bert4rec}. Table~\ref{tab:datasets} summarizes the dataset statistics.

\begin{table}[!htbp]
    \centering
    \footnotesize
    \renewcommand{\arraystretch}{0.8} 
    \setlength{\tabcolsep}{4pt}          
    \caption{\(AUC_{NDCG@10}\) (mean \(\pm\) std across seeds) across different model architectures on $\mathcal{D}_{\mathrm{valid}}$.}
    \label{tab:ndcg_combined}
    \begin{adjustbox}{max width=\textwidth}
    \begin{tabular}{llcccc}
    \toprule
    Method & Submethod & ML-1M & ML-10M & KuaiRand-1K & KuaiRand-1K-Long \\
    \midrule
    \multicolumn{6}{l}{\textbf{\textit{SASRec}}} \\
    Random & -- & $0.679 \pm 0.012$ & $0.718 \pm 0.036$ & $0.90 \pm 0.10$ & $0.88 \pm 0.12$ \\
    Recency & -- & $\underline{0.969 \pm 0.025}$ & $\mathbf{1.003 \pm 0.016}$ & $0.943 \pm 0.034$ & $0.992 \pm 0.063$ \\
    \multirow{4}{*}{\shortstack[l]{Ours\\(XAI-Driven)}} & \cellcolor{lightmaroon}Integrated Gradient & \cellcolor{lightmaroon}$0.961 \pm 0.041$ & \cellcolor{lightmaroon}$0.976 \pm 0.024$ & \cellcolor{lightmaroon}$\mathbf{1.175 \pm 0.056}$ & \cellcolor{lightmaroon}$\underline{1.10 \pm 0.18}$ \\
     & \cellcolor{lightmaroon}GradientSHAP & \cellcolor{lightmaroon}$\mathbf{0.972 \pm 0.017}$ & \cellcolor{lightmaroon}$\underline{0.987 \pm 0.013}$ & \cellcolor{lightmaroon}$\underline{1.146 \pm 0.047}$ & \cellcolor{lightmaroon}$1.10 \pm 0.20$ \\
     & \cellcolor{lightmaroon}Attention Tracing & \cellcolor{lightmaroon}$0.781 \pm 0.039$ & \cellcolor{lightmaroon}$0.822 \pm 0.026$ & \cellcolor{lightmaroon}$0.928 \pm 0.037$ & \cellcolor{lightmaroon}$1.06 \pm 0.14$ \\
     & \cellcolor{lightmaroon}Grad-SAM & \cellcolor{lightmaroon}$0.902 \pm 0.037$ & \cellcolor{lightmaroon}$0.926 \pm 0.050$ & \cellcolor{lightmaroon}$1.049 \pm 0.089$ & \cellcolor{lightmaroon}$\mathbf{1.12 \pm 0.15}$ \\
    \addlinespace[2pt]
    \multicolumn{6}{l}{\textbf{\textit{BERT4Rec}}} \\
    Random & -- & $0.568 \pm 0.036$ & $0.730 \pm 0.038$ & $1.101 \pm 0.067$ & $1.15 \pm 0.15$ \\
    Recency & -- & $\mathbf{0.873 \pm 0.065}$ & $0.884 \pm 0.046$ & $1.06 \pm 0.10$ & $1.14 \pm 0.16$ \\
    \multirow{4}{*}{\shortstack[l]{Ours\\(XAI-Driven)}} & \cellcolor{lightmaroon}Integrated Gradient & \cellcolor{lightmaroon}$0.851 \pm 0.066$ & \cellcolor{lightmaroon}$\underline{0.903 \pm 0.021}$ & \cellcolor{lightmaroon}$1.36 \pm 0.41$ & \cellcolor{lightmaroon}$\mathbf{1.49 \pm 0.48}$ \\
     & \cellcolor{lightmaroon}GradientSHAP & \cellcolor{lightmaroon}$0.813 \pm 0.054$ & \cellcolor{lightmaroon}$0.897 \pm 0.038$ & \cellcolor{lightmaroon}$\underline{1.46 \pm 0.41}$ & \cellcolor{lightmaroon}$\underline{1.38 \pm 0.28}$ \\
     & \cellcolor{lightmaroon}Attention Tracing & \cellcolor{lightmaroon}$\underline{0.860 \pm 0.053}$ & \cellcolor{lightmaroon}$\mathbf{0.926 \pm 0.033}$ & \cellcolor{lightmaroon}$0.92 \pm 0.17$ & \cellcolor{lightmaroon}$0.94 \pm 0.19$ \\
     & \cellcolor{lightmaroon}Grad-SAM & \cellcolor{lightmaroon}$0.802 \pm 0.038$ & \cellcolor{lightmaroon}$0.802 \pm 0.016$ & \cellcolor{lightmaroon}$\mathbf{1.83 \pm 0.74}$ & \cellcolor{lightmaroon}$1.20 \pm 0.33$ \\
    \addlinespace[2pt]
    \multicolumn{6}{l}{\textbf{\textit{Mamba}}} \\
    Random & -- & $0.637 \pm 0.010$ & $0.672 \pm 0.030$ & $0.87 \pm 0.13$ & $0.89 \pm 0.10$ \\
    Recency & -- & $\mathbf{0.966 \pm 0.020}$ & $\underline{0.983 \pm 0.016}$ & $0.96 \pm 0.12$ & $0.94 \pm 0.12$ \\
    \multirow{2}{*}{\shortstack[l]{Ours\\(XAI-Driven)}} & \cellcolor{lightmaroon}Integrated Gradient & \cellcolor{lightmaroon}$\underline{0.954 \pm 0.014}$ & \cellcolor{lightmaroon}$\mathbf{0.985 \pm 0.022}$ & \cellcolor{lightmaroon}$\mathbf{1.27 \pm 0.16}$ & \cellcolor{lightmaroon}$\underline{1.01 \pm 0.12}$ \\
     & \cellcolor{lightmaroon}GradientSHAP & \cellcolor{lightmaroon}$0.950 \pm 0.013$ & \cellcolor{lightmaroon}$0.980 \pm 0.029$ & \cellcolor{lightmaroon}$\underline{1.17 \pm 0.16}$ & \cellcolor{lightmaroon}$\mathbf{1.04 \pm 0.10}$ \\
    \addlinespace[2pt]
    \multicolumn{6}{l}{\textbf{\textit{LONGER}}} \\
    Random & -- & $0.580 \pm 0.032$ & $0.561 \pm 0.059$ & $0.80 \pm 0.11$ & $0.772 \pm 0.079$ \\
    Recency & -- & $\mathbf{0.845 \pm 0.045}$ & $\underline{0.783 \pm 0.092}$ & $0.83 \pm 0.22$ & $0.84 \pm 0.10$ \\
    \multirow{2}{*}{\shortstack[l]{Ours\\(XAI-Driven)}} & \cellcolor{lightmaroon}Integrated Gradient & \cellcolor{lightmaroon}$\underline{0.774 \pm 0.056}$ & \cellcolor{lightmaroon}$\mathbf{0.796 \pm 0.050}$ & \cellcolor{lightmaroon}$\underline{1.03 \pm 0.12}$ & \cellcolor{lightmaroon}$\mathbf{1.09 \pm 0.12}$ \\
     & \cellcolor{lightmaroon}GradientSHAP & \cellcolor{lightmaroon}$0.747 \pm 0.052$ & \cellcolor{lightmaroon}$0.743 \pm 0.065$ & \cellcolor{lightmaroon}$\mathbf{1.09 \pm 0.12}$ & \cellcolor{lightmaroon}$\underline{1.012 \pm 0.056}$ \\
    \bottomrule
    \end{tabular}
    \end{adjustbox}
\end{table}

 \subsection{Baseline Methods}
\vspace{-2mm}

  We compare XAI-based distillation against two model-agnostic baselines that select retained subsequences without attribution signals. \textbf{Random distillation} selects $K$ positions uniformly at random as a reference baseline. \textbf{Recency-based distillation} retains the $K$ most recent interactions, representing the standard heuristic that recent user history dominates next-item prediction. Both baselines are evaluated under identical settings to isolate the specific impact of attribution-guided selection.

To evaluate generalization across architectural paradigms, we conduct our study across four distinct sequence backbones: \textbf{SASRec}~\citep{kang2018sasrec} (autoregressive Transformer), \textbf{BERT4Rec}~\citep{sun2019bert4rec} (bidirectional Transformer), \textbf{Mamba}~\citep{gu2023mamba} (state-space model), and \textbf{LONGER}~\citep{longer} (linear-complexity architecture). All distillation strategies are evaluated across these architectures under an identical protocol. For non-attention models (Mamba and LONGER), Attention Tracing and Grad-SAM are excluded as they rely on explicit attention matrices; Integrated Gradients and GradientSHAP are evaluated across all four architectures.

To benchmark against models designed for extended contexts, we compare against full-sequence ($K\!=\!L$) executions of \textbf{GLASS}~\citep{cao2026glass}, a coarse-to-fine generative recommender utilizing semantic ID search, and our linear-complexity \textbf{LONGER} backbone~\citep{longer}. It is important to note that \textbf{LONGER} is also considered as one of the backbones to show the impact of sequence distillation. However, \textbf{LONGER}'s performance at $K=L$ is one of the fair baseline for our work. Comparing against these purpose-built baselines demonstrates whether attribution-guided distillation on standard backbones can match or exceed specialized long-sequence paradigms while maintaining favorable inference footprints (GFLOPs).

\subsection{Evaluation Strategy}\label{eval}
\vspace{-2mm}
We evaluate all distillation methods by measuring downstream ranking performance as a function of the retained sequence length $K$. Across datasets, we across several $K$ values: for MovieLens (ML)-1M ($L=200$), $K \in \{200, 175, 150, 125, 100, 50, 25, 12, 6, 3, 1\}$; for ML-10M ($L=500$) and KuaiRand-1K ($L=500$), $K \in \{500, 400, 300, 200, 100, 75, 50, 25, 12, 6, 3, 1\}$; and for KuaiRand-1K-Long ($L=1000$), $K \in \{1000, 900, 800, 700, 600, 500, 400, 300, 200, 100, 75, 50, 25, 12, 6, 3, 1\}$ to probe history distillation across extended sequence horizons. In every configuration, $K=L$ corresponds to the full, undistilled sequence serving as the normalization baseline in Equation~\ref{normalized_eq}. Crucially, to prevent data leakage and evaluate realistic deployment, the search grid over $K$ is swept strictly on the validation set $\mathcal{D}_{\mathrm{valid}}$ to identify the most efficient operating point $K^*$; all final reported operating points, performance curves, and compute trade-offs are subsequently evaluated on the held-out test set $\mathcal{D}_{\mathrm{test}}$.

All models are implemented in PyTorch and trained using the Adam optimizer ($\beta_1=0.9$, $\beta_2=0.98$) with an initial learning rate of $1\text{e-}3$. For each architecture-dataset configuration, experiments are conducted across three independent random seeds, with all reported metrics presenting the mean $\pm$ standard deviation to ensure statistical reliability and verify that observed performance margins are not artifacts of seed initialization. Additional implementation specifications, architectural hyperparameters, and training protocols are detailed in Appendix~\ref{train_detail}.

\section{Results and Discussions}
\vspace{-3mm}
We organize our findings around three questions: does XAI-driven distillation improve ranking quality over attribution-agnostic baselines across the full  compression spectrum (Section~\ref{sec:res-effectiveness}); does this advantage translate into a favorable performance-efficiency trade-off at each  method's best operating point (Section~\ref{sec:res-efficiency}); and does the benefit of attribution-guided distillation persist under varying model  capacity and noisy input sequences (Section~\ref{sec:res-ablation}).

\subsection{Effectiveness of XAI-driven Sequence Distillation} \label{sec:res-effectiveness}

Table~\ref{tab:ndcg_combined} reports \(\mathrm{AUC}_{\mathrm{NDCG}@10}\), as defined in Equation~\ref{eq:auc_metric} and computed from the compression trajectories in Appendix Figure~\ref{fig:app_arch_ndcg10}. This holistic metric quantifies how effectively each distillation strategy preserves ranking utility across the entire spectrum of sequence budgets, rather than evaluating performance at an isolated retained length \(K\). While \({AUC}_m\) scales with the evaluation grid cardinality \(|{K}|\), empirical stability is achieved once \(|{K}| > 4\) (Figure~\ref{fig:auc_sensitivity}); our protocol adheres to this threshold across all benchmarks as established in Section~\ref{eval}. Aggregated across all backbone architectures (SASRec, BERT4Rec, Mamba, LONGER) and datasets (ML-1M, ML-10M, KuaiRand-1K, KuaiRand-1K-Long), XAI-driven distillation delivers a consistent net advantage over the Recency baseline. Averaged over all evaluated settings, Integrated Gradients and GradientSHAP improve over Recency's median \(\mathrm{AUC}_{\mathrm{NDCG}@10}\) by \(+11.2\%\) and \(+9.1\%\), respectively, while Grad-SAM improves by \(+4.5\%\). \begin{wrapfigure}{r}{0.45\textwidth}
      \centering
      \includegraphics[width=0.4\textwidth]{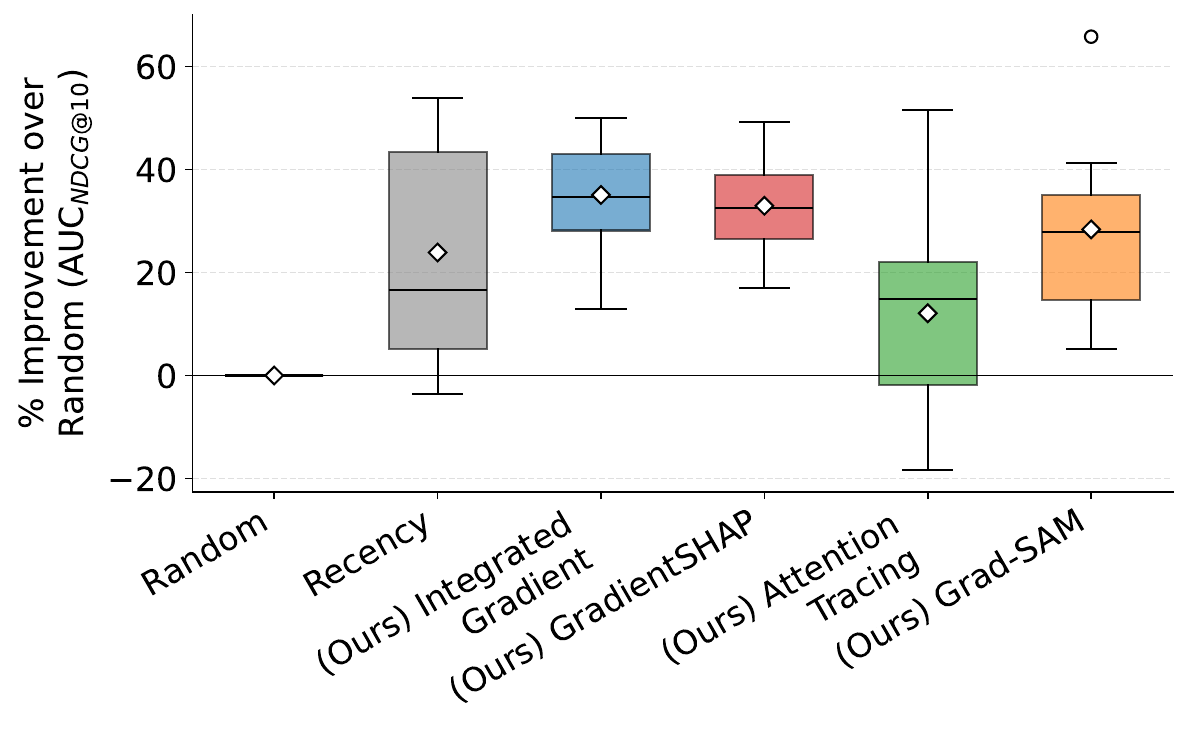}
      \vspace{-5mm}
      \caption{Percentage improvement in \(AUC_{NDCG@10}\) over the \textbf{Random} baseline.}
      \label{fig:method_advantage_box}
\end{wrapfigure} Attention Tracing is the exception, trailing Recency by \(11.8\%\) on average. This indicates that raw attention weights are not a reliable indicator of sequence position importance, a limitation consistent with prior literature~\citep{attention_not_working1, attention_not_working2, attention_not_working3, attention_not_working4,pandey2026cars}. This pattern is corroborated across dataset-architecture combinations in Figure~\ref{fig:method_advantage_box}, where Integrated Gradients and GradientSHAP display the most consistent improvements over $Random$ method while Attention Tracing is the only method whose performance distribution dips into negative territory. Crucially, this performance edge is directional and consistent across all settings: Integrated Gradients and GradientSHAP exceed Recency in all eight KuaiRand configurations (4 backbones \(\times\) 2 sequence lengths) in Table~\ref{tab:ndcg_combined}. This strict sign-consistency demonstrates that the relative advantage is robust to per-cell variance.

\begin{figure}[H]
      \centering
      \includegraphics[width=\textwidth]{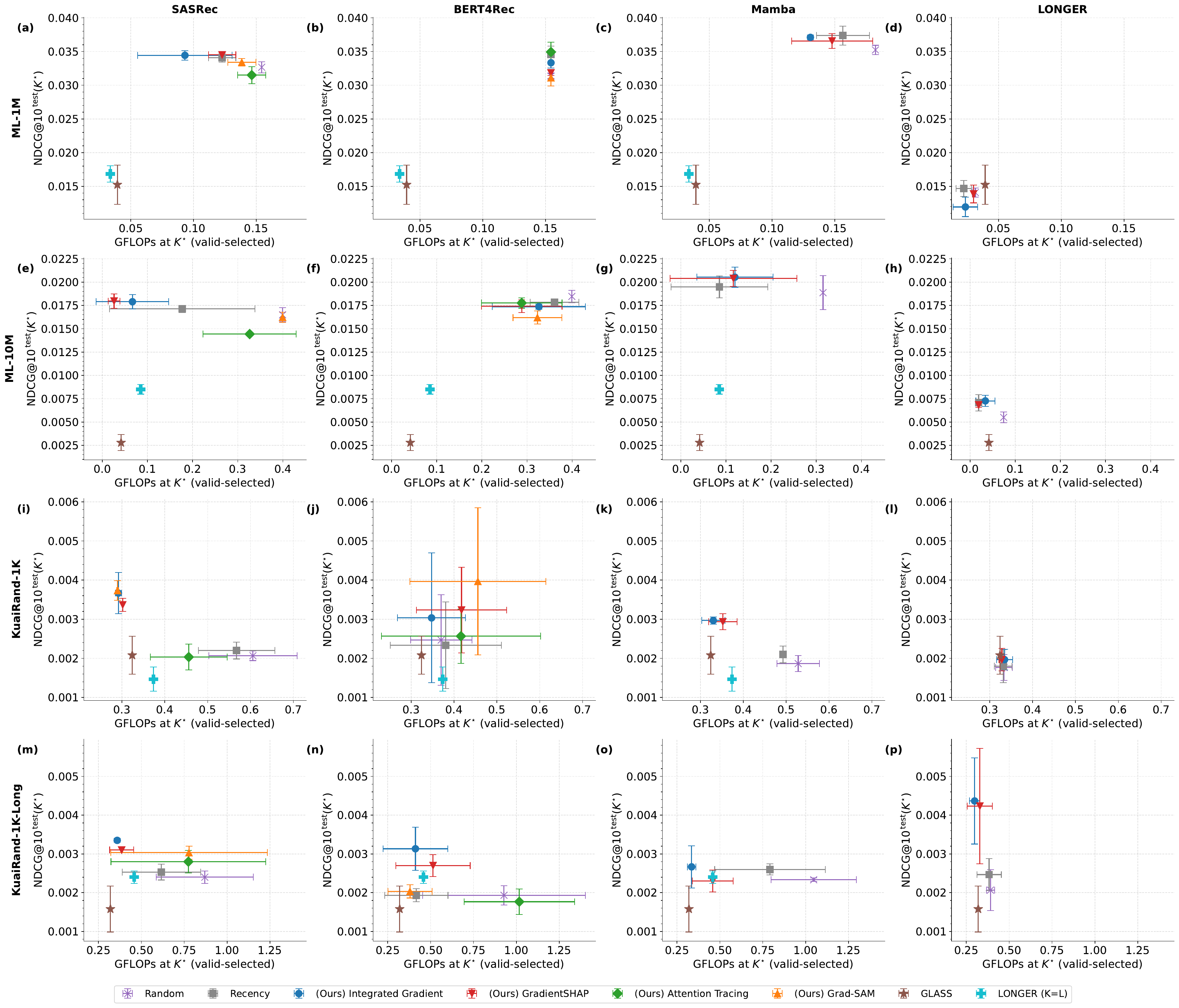}
      \vspace{-7mm}
      \caption{\textbf{Peak NDCG@10 vs.\  GFLOPs} across model architectures (columns) and datasets (rows). Points closer to the upper-left (higher NDCG@10 at lower GFLOPs) represent a \textbf{better performance-efficiency trade-off}.}
      \label{fig:pareto_sasrec_ndcg}
  \end{figure}
  
This advantage exhibits systematic shifts across domains: on ML-1M, Recency serves as a strong baseline due to pronounced recency bias (Figure \ref{fig:recnecy_bias}), leading all attribution methods across backbones by $3.1\%\text{--}10.1\%$ in mean $\mathrm{AUC}_{\mathrm{NDCG}@10}$. On ML-10M, however, this margin narrows sharply as Integrated Gradients (IG) draws level with Recency ($+0.2\%$), and any remaining Recency edge disappears under compute-constrained Pareto trade-offs and noise-robustness evaluations. Furthermore, evaluating standard, linear, and sparse attention variants of SASRec reveals near-identical performance--compute curves, confirming that sequence distillation is governed primarily by dataset interaction dynamics rather than the underlying attention architecture (Figure~\ref{fig:sasrec_ig_attn_variants}). Conversely, KuaiRand-1K exhibits weak temporal locality due to fragmented short-video engagement signals, allowing IG and GradientSHAP to surpass Recency by $27.6\%$ and $28.4\%$. This attribution advantage persists on KuaiRand-1K-LONG with improvements between $15.8\%\text{--}19.7\%$, demonstrating that these efficiency gains stem from underlying engagement dynamics rather than sequence length alone. Similar analysis for $AUC_{HR@10}$ is presented in Table~\ref{tab:hr_combined}. 
  
\subsection{XAI-driven Sequence Distillation offer Higher Efficiency} \label{sec:res-efficiency}
\vspace{-2mm}

While $\mathrm{AUC}_{\mathrm{NDCG}@10}$ provides a holistic summary across sequence budgets $K$, it does not capture the operational efficiency attained at a deployment-ready operating point. Figure~\ref{fig:pareto_sasrec_ndcg} complements Table~\ref{tab:ndcg_combined} by plotting test $\mathrm{NDCG}@10$ at each method's validation-selected $K^\star$ against inference compute (GFLOPs), where top-left positions indicate optimal Pareto trade-offs.

Across datasets, gradient-based attribution methods match or Pareto-dominate heuristic Recency truncation at these operating points on all but two of the 16 dataset$\times$architecture combinations, where Recency retains an edge (both on ML-1M). For SASRec on KuaiRand-1K (panel i), IG and GradientSHAP reach $\mathrm{NDCG}@10$ of $0.0037$ and $0.0034$ at approximately $0.30$ GFLOPs, whereas Recency requires $0.57$ GFLOPs to achieve $0.0022$. KuaiRand-1K-LONG (panel m) exhibits a matching pattern: IG and GradientSHAP reach $0.0033$ and $0.0031$ at roughly $0.37$ GFLOPs, while Recency trails at $0.0025$ with $0.62$ GFLOPs. Similarly, on ML-10M (panel e), GradientSHAP and IG exceed Recency's peak ranking quality ($0.0180$ and $0.0179$ vs.\ $0.0171$) while sharply cutting serving compute ($0.03$ and $0.07$ vs.\ $0.18$ GFLOPs).

On ML-1M, attribution-guided distillation across all architectures either closely matches or surpasses recency-based truncation along the Pareto frontier. As illustrated across panels (f)--(h), (j)--(l), and (n)--(p), this Pareto dominance extends consistently to BERT4Rec, Mamba, and LONGER backbones. While full-length \textbf{LONGER} ($K=L$) and generative compression (\textbf{GLASS}) incur low GFLOPs at inference, neither achieves the peak ranking accuracy of attribution-distilled models. Consequently, executing sequence pruning offline substantially curtails online serving compute while preserving or exceeding full-sequence ranking fidelity.

\noindent\textbf{Performance at the fixed compute budget:} Beyond evaluating methods at their individual optimal operating points ($K^\star$), practical deployments frequently require serving under strictly predetermined inference budgets. To isolate selection quality from operating-point bias, we further evaluate all distillation methods under identical, fixed sequence lengths $K \in \{100, 50, 25, 12, 6\}$ across all architectures and seeds in Appendix \ref{statstical_analysis}. Across both metrics (HR@10 and NDCG@10), a Friedman test \citep{fridman} decisively confirms non-equivalent performance ($p < 10^{-60}$), with IG securing the best overall rank across benchmarks. Notably, gradient-based methods achieves the top rank on every dataset except ML-1M, where Recency outperforms it due to pronounced temporal recency bias. Full per-budget performance breakdowns, critical difference intervals, and detailed Friedman rank tables are provided in Appendix~\ref{statstical_analysis}.

\subsection{Ablation Study} \label{sec:res-ablation}
\vspace{-2mm}
\subsubsection{Impact of XAI-driven sequence distillation for noisy input}
\vspace{-2mm}
To assess whether attribution-guided distillation remains beneficial when the interaction history is corrupted, we inject random noise into test sequences before computing attributions and evaluating downstream performance. For a noise level $\eta \in \{10\%, 30\%, 50\%\}$, we independently sample $\lceil \eta \cdot L \rceil$ positions per sequence without replacement and replace each sampled item with a uniform sample from the full vocabulary. This corruption is applied exclusively to $\mathcal{D}_{\text{test}}$, leaving training and validation sequences unperturbed. Attribution scores are then recomputed on each corrupted sequence to guide distillation to varying target lengths $K$, following the protocol in Section~\ref{method_description}.
\vspace{-3mm}
\begin{figure}[H]
      \centering
      \includegraphics[width=\textwidth]{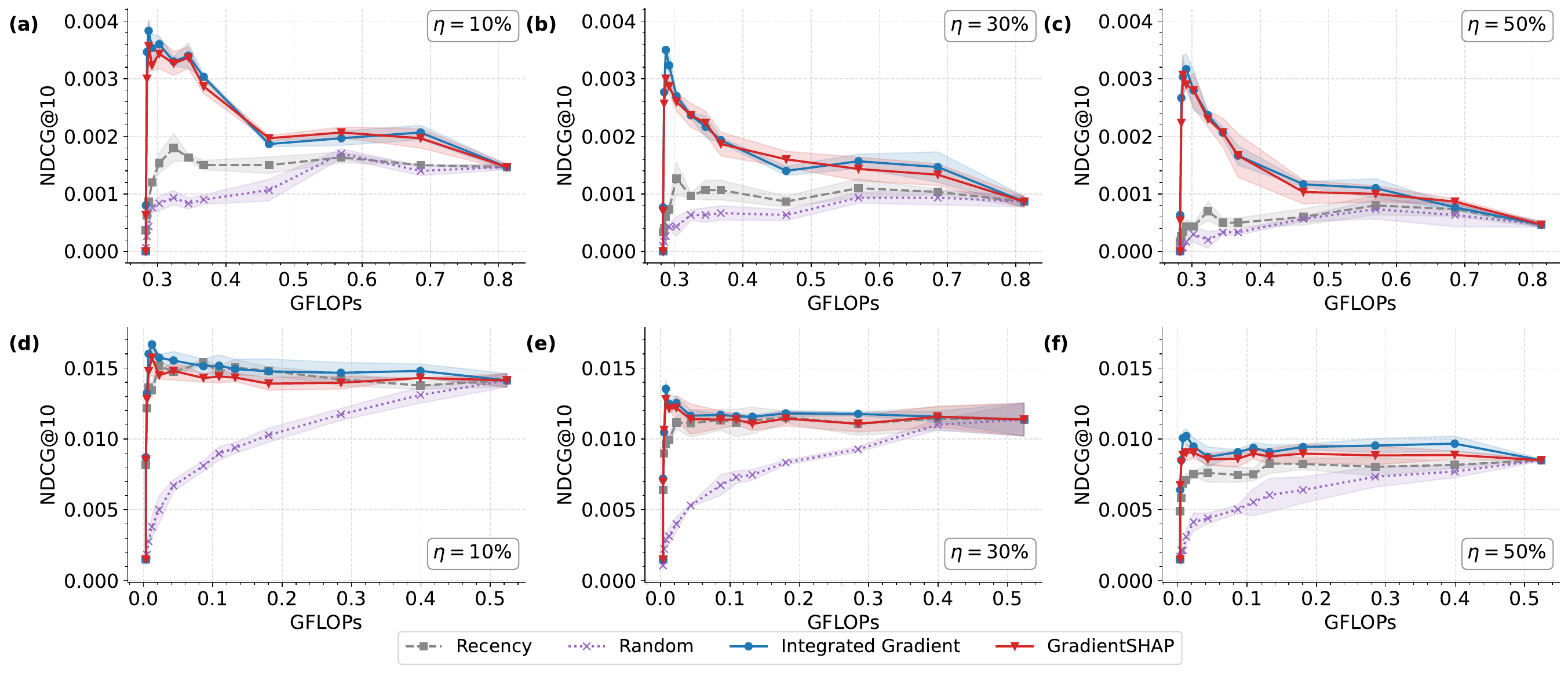}
      \vspace{-8mm}
      \caption{NDCG@10 as a function of GFLOPs for SASRec on KuaiRand-1K (a-c) and ML-1M (d-f) under random item noise injection at three noise levels ($\eta \in \{10\%, 30\%, 50\%\}$). The \textbf{rightmost point} in each graph represents the performance with \textbf{no distillation}.}
      \label{fig:noise_ablation}
  \end{figure}
\vspace{-5mm}
Figure~\ref{fig:noise_ablation} reports NDCG@10 against GFLOPs for SASRec under this setup, with KuaiRand-1K in the top row (panels a--c) and ML-10M in the bottom row (panels d--f). On KuaiRand-1K, XAI-guided distillation provides a substantially larger benefit than Recency across all noise levels: Integrated Gradients and GradientSHAP sustain a peak between $0.003$ and $0.004$ NDCG@10, while Recency remains well below this plateau. On ML-1M, while clean-input benchmarks show comparable performance between Recency and attribution-based methods (Table~\ref{tab:ndcg_combined}, Figure~\ref{fig:pareto_sasrec_ndcg}), XAI-guided methods demonstrate superior resilience as sequence noise increases. Crucially, across both datasets, explainability-driven distillation improves performance over the uncompressed sequence ($K=L$), showcasing a clear denoising effect that filters out corrupted interactions. Averaged across all noise levels and both datasets, Integrated Gradients and GradientSHAP outperform Recency by 37\% and 33\%, respectively, in $\text{AUC}_{\text{NDCG@10}}$, demonstrating that attribution-guided selection remains the more reliable distillation strategy under input noise.\begin{wrapfigure}{r}{0.45\textwidth}
      \centering
      \includegraphics[width=0.43\textwidth]{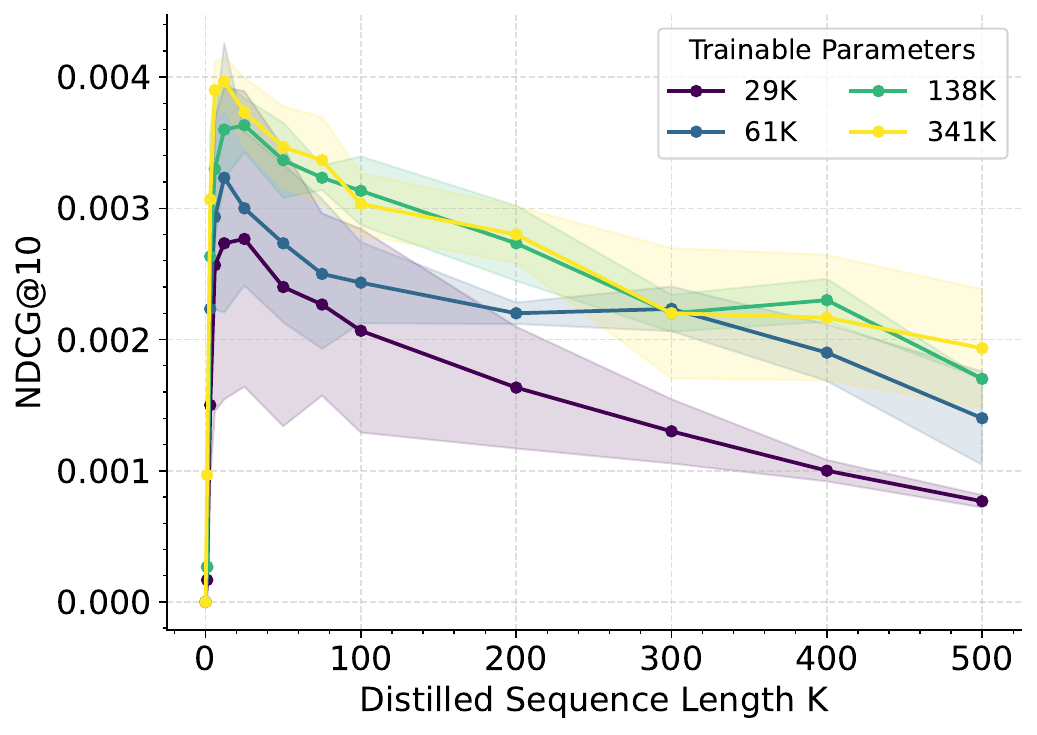}
      \vspace{-3mm}
      \caption{NDCG@10 as a function of distilled sequence length $K$ for SASRec on KuaiRand-1K, using Integrated Gradients across four model sizes. Shaded regions denote standard deviation across three seeds.}
      \label{fig:model_size_ablation}
  \end{wrapfigure} To simulate real-world noise from accidental clicks or transient exploration, we corrupt the most recent interactions in test sequences (Figure~\ref{fig:recency_noise_ablation}). On KuaiRand-1K, Integrated Gradients and GradientSHAP remain robust across all noise levels $\eta \in \{10\%, 30\%, 50\%\}$, whereas recency truncation degrades sharply below random selection. On ML-10M, while all methods decline due to strong temporal reliance (Figure~\ref{fig:recnecy_bias}), attribution-guided methods widen their relative advantage. Across both datasets and noise ratios, Integrated Gradients and GradientSHAP achieve nearly twice the $\text{AUC}_{\text{NDCG@10}}$ of Recency, demonstrating that XAI-guided distillation provides critical resilience when recent interactions are unreliable.

\vspace{-2mm}
\subsubsection{Effect of model size on sequence distillation}
\vspace{-2mm}

    Figure~\ref{fig:model_size_ablation} characterizes the interaction between model capacity and XAI-guided distillation for SASRec on KuaiRand-1K, sweeping four parameter configurations and reporting raw NDCG@10 as a function of the distilled sequence length $K$. Capacity governs both full-sequence performance and the achievable peak after distillation: the two largest models (138K and 341K) dominate throughout the compression spectrum, attaining both the highest full-sequence NDCG@10 and the highest peaks once Integrated Gradients selects a small, high-value subset of positions. The compact models (29K and 61K parameters) lag behind at all sequence lengths $K$, with the performance gap expanding under higher compression. This behavior stems from under-parameterization, where restricted model capacity limits representations learned from the full history. Although distillation provides relative gains over the undistilled baseline, its absolute ceiling remains bounded by the model's base capacity. Thus, XAI-guided distillation relies on sufficient architectural capacity to maximize efficiency gains. 

\vspace{-3mm}
\section{Conclusion}
\vspace{-3mm}

We present explainability-guided sequence distillation, a retraining-free framework leveraging post-hoc attribution to distill user interaction histories and drastically reduce online inference GFLOPs. Across four backbone architectures and four dataset configurations, our approach consistently outperforms standard recency truncation, matching or exceeding it even under heavy recency bias when evaluating budget-matched compute. Comprehensive ablations show these gains persist and amplify under sequence corruption. Ultimately, our work repurposes explainability from a passive diagnostic into an active driver of production online inference efficiency across diverse sequential recommendation models. Future directions include live deployment and extension to broader sequential domains.

\noindent{}\textbf{Limitations.} Our empirical findings demonstrate that gradient-based attribution provides a reliable signal for sequence distillation, substantially curtailing online serving GFLOPs while matching or surpassing full-sequence ranking fidelity. However, a primary operational constraint is the computational overhead required to generate these attribution scores offline. As detailed in Table~\ref{tab:xai_gflops_sasrec}, computing gradient-based importances at sequence length $L$ demands considerable per-example GFLOPs, most notably for multi-step methods such as Integrated Gradients, as well as GradientSHAP and Grad-SAM. Although this cost is strictly isolated to periodic offline precomputation and never directly impacts live serving latency, the scaling overhead can be nontrivial for extremely high-throughput catalog pipelines. Crucially, this study establishes that gradient attribution acts as an effective signal for sequence relevance, laying a foundational reference for future work to develop lightweight, single-pass explainability estimators that approximate these attribution scores without requiring iterative integration steps or backward passes.
  

\subsection*{AI use statement}

In this work, we used generative AI tools for refining the textual draft and generating visualization code for figures. We take full responsibility for the final content of this work, including all text, claims, and artifacts produced with the aid of generative AI.

\subsubsection*{Reproducibility Statement}
To ensure the reproducibility of our findings, we have provided comprehensive details of our experimental setup, architectural parameters, and evaluation metrics in Section~\ref{method} and Appendix~\ref{app:model_detail}. Furthermore, the complete codebase, including data preprocessing scripts, backbone model implementations, post-hoc attribution pipelines, and evaluation routines, is included as supplementary material and will be made fully open-source upon paper acceptance.



\bibliography{nips_2026}
\bibliographystyle{plainnat}

\appendix
\section{Model Details} \label{app:model_detail}

\paragraph{Prediction task and output dimension.}
All models are trained for next-item prediction, where the target label is the index of the subsequent interacted item. The prediction head scores the entire post-preprocessing item catalog, yielding an output dimension of $|\mathcal{V}|+1$ (corresponding to the \textbf{Items} column in Table~\ref{tab:datasets} plus a reserved padding index masked during ranking). For the KuaiRand benchmarks, this yields $1{,}101{,}222+1$ classes on KuaiRand-1K and $1{,}068{,}905+1$ on KuaiRand-1K-Long.

\paragraph{Candidate set protocol (Tables~\ref{tab:datasets}--\ref{app_tab:num_parameters} vs.\ Section~\ref{train_detail}).}
All reported test metrics reflect \emph{full-catalog} evaluation, wherein the ground-truth item is ranked against all $|\mathcal{V}|$ candidates without negative sampling. The $100$ sampled negatives referenced in Section~\ref{train_detail} are utilized strictly to accelerate training and validation sweeps. Because the underlying backbone and inner-product scoring function ($\mathbf{h}^\top \mathbf{e}_v$) remain unchanged, the two evaluation regimes differ solely in candidate-set cardinality rather than model architecture or target formulation.

\paragraph{Content feature encoding and parameter scaling.}
On KuaiRand, items are \emph{content-encoded} rather than represented via discrete ID lookup tables: each item is parameterized as a $61$-dimensional feature vector (\texttt{video\_type} one-hot ($3$) concatenated with multi-hot tags ($58$)) and mapped into the latent space via a learned $\mathrm{Linear}(61 \to E)$ projection. Similarly, user profiles are encoded via $\mathrm{Linear}(15 \to E)$, augmented with $9$-dimensional per-step interaction features and a $\log(1+\Delta t)$ temporal projection. Because the architecture omits per-item and per-user embedding tables, the $>10^6$-item catalog size reported in Table~\ref{tab:datasets} does not contribute to the parameter counts in Table~\ref{app_tab:num_parameters}. Instead, trainable parameters scale exclusively with feature dimensionality, latent width $E$, and maximum sequence length $L$, explaining why KuaiRand-1K and KuaiRand-1K-Long require only $341\text{K}$ and $405\text{K}$ trainable parameters despite million-scale catalogs.

\paragraph{Parameter-counting scope.}
Frozen content-feature matrices are registered as non-trainable buffers and excluded from the totals in Table~\ref{app_tab:num_parameters}; only learnable projection layers, sequence backbone weights, and output-head parameters are tallied. Consequently, the exact $+64\text{K}$ parameter increase between KuaiRand-1K ($341\text{K}$) and KuaiRand-1K-Long ($405\text{K}$) stems entirely from the learned positional embeddings scaling with sequence length ($L: 500 \to 1000$, adding $500 \times E$ parameters at $E=128$).

\paragraph{Accounting for full-catalog scoring in GFLOPs.}
Full-catalog evaluation cost is explicitly incorporated into our GFLOPs measurements as the per-query inner product across all catalog items, requiring $2E(|\mathcal{V}|+1)$ FLOPs. Because the projected item representation table (generated by applying \texttt{item\_proj} across static item features) can be cached offline and amortized across inference requests, and the final $\operatorname{argsort}$ ranking operation consists of non-arithmetic comparisons rather than floating-point multiply--adds, neither is included in per-query inference FLOPs.

\begin{figure}[H]
      \centering
      \includegraphics[width=\textwidth]{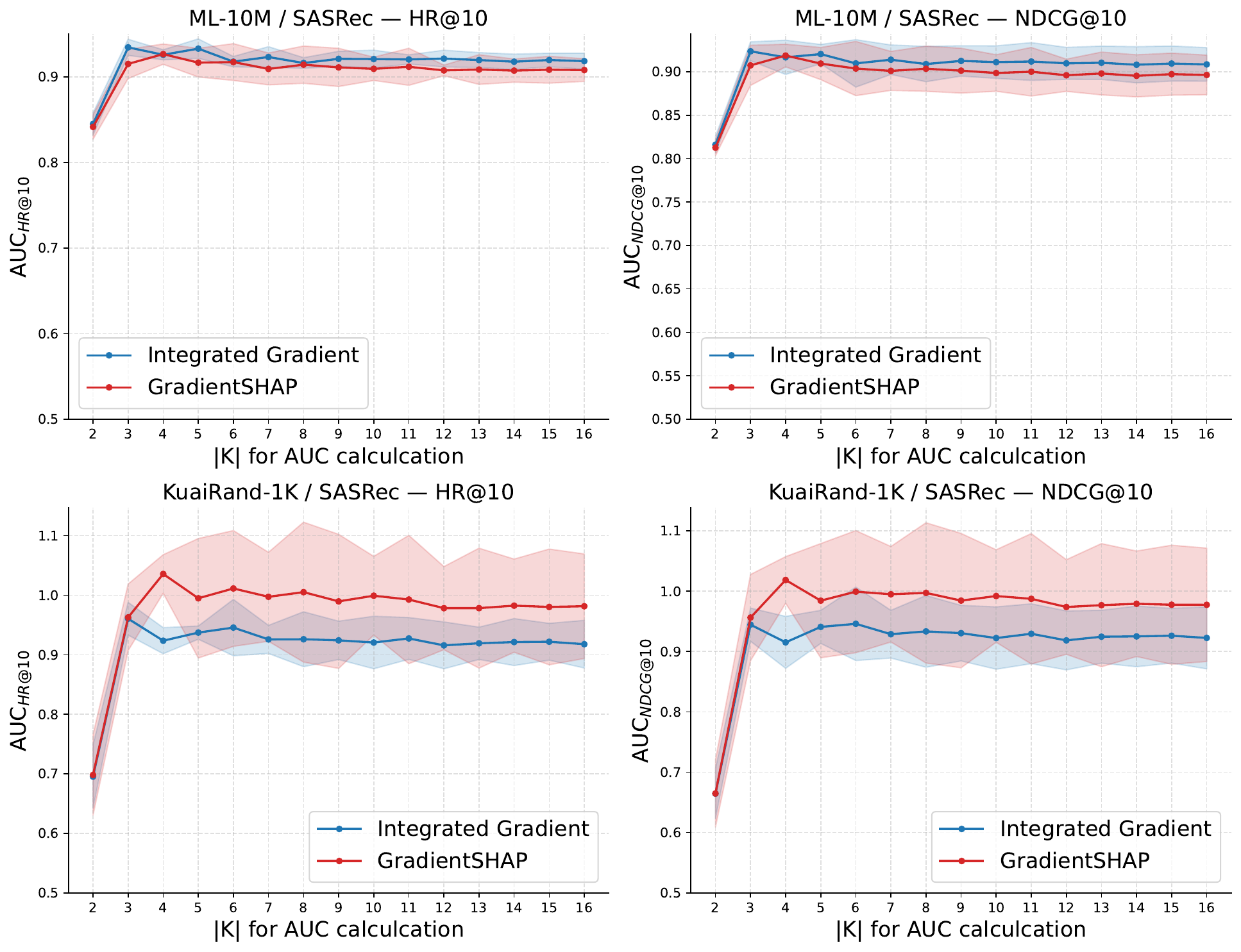}
      \caption{Sensitivity of $AUC_m$ with respect to the cardinality of K i.e., $|$K$|$.}
      \label{fig:auc_sensitivity}
  \end{figure}
  
\begin{table}[!htbp]
      \centering
      \caption{Dataset statistics.}
      \label{tab:datasets}
      \footnotesize
      \setlength{\tabcolsep}{5pt}
      \renewcommand{\arraystretch}{1}
      \begin{tabular}{lrrrrrrrr}
      \toprule
      \textbf{Dataset} & $L$ & \textbf{Users} & \textbf{Items} & \textbf{Sparsity} & $|\mathcal{D}_{train}|$ & $|\mathcal{D}_{valid}|$ & $|\mathcal{D}_{test}|$ \\
      \midrule
      KuaiRand-1K  & 500  & 922   & 1,101,222 & 99.75\% & 28,267 & 6,149 & 6,553 \\
      KuaiRand-1K-LONG  & 1000 & 786   & 1,068,905   & 99.72\%     & 22,314   & 4,858    & 5,205   \\
      MovieLens-1M & 200  & 1,578 & 3,883,   & 89.31\% & 17,731 & 2,813 & 3,080 \\
      MovieLens-10M & 500  & 3,659 & 10,583 & 91.99\% & 17,973 & 4,672 & 4,728 \\
      \bottomrule
      \end{tabular}
  \end{table}

\subsection{Training Details}\label{train_detail}
All models were trained using NVIDIA A10G GPUs for a total of 200 epochs. Table~\ref{app_tab:num_parameters} details the number of trainable parameters for the SASRec and BERT4Rec architectures across the KuaiRand datasets.

\begin{table*}[!htbp]
    \centering
    \caption{Number of trainable parameters for SASRec, BERT4Rec, Mamba, and LONGER architectures across datasets.}
    \label{app_tab:num_parameters}
    \begin{adjustbox}{max width=\textwidth}
    \begin{tabular}{llcccc}
    \toprule
    \textbf{Architecture} & \textbf{Variant} & \textbf{KuaiRand-1k} & \textbf{KuaiRand-1k-LONG} & \textbf{MovieLens-1M} & \textbf{MovieLens-10M} \\
    \midrule
    \multirow{2}{*}{Transformer}
        & SASRec   & 341K & 405K & 1.56M & 2.15M \\
        & BERT4Rec & 341K & 405K & 1.56M & 2.15M \\
    \midrule
    \multirow{2}{*}{Non-Transformer}
        & Mamba    & 442K & 506K & 1.55M & 2.14M \\
        & LONGER   & 524K & 588K & 1.74M & 2.34M \\
    \bottomrule
    \end{tabular}
    \end{adjustbox}
\end{table*}

\subsection{Integrated Gradient and GradientSHAP hyperparameters:}
\label{app:ig_grad}
For Integrated Gradients, we integrate over $n_{steps}=20$ interpolation steps from a zero
  baseline (i.e., the all-zero item-embedding tensor), matching Captum's default Riemann-sum approximation. For GradientSHAP, we draw
  $n_{samples}=2$ perturbed baselines per input from a Gaussian noise distribution centered at zero with standard deviation $0.001$ (four
  such noise baselines are pre-sampled and shared across the batch), and apply a smoothing noise level of $stdevs=0.05$ around each
  sampled baseline, following Captum's implementation of the expected-gradients formulation. Both methods attribute importance with
  respect to the single top-predicted next item ($top_k=1$) at each position, rather than aggregating over multiple candidate targets.
  These settings are held fixed across all architectures (SASRec, BERT4Rec, Mamba, LONGER), attention variants, and datasets to ensure a
  controlled comparison; we did not perform hyperparameter tuning for either attribution method, as our goal is to evaluate off-the-shelf
  explainability signals rather than to optimize attribution quality itself.

  \begin{table}[t]
  \centering
  \caption{Per-example GFLOPs to compute each importance score for SASRec.}
  \label{tab:xai_gflops_sasrec}
  \begin{tabular}{lrrrr}
  \toprule
  Dataset ($L$) & Integrated Gradient & GradientSHAP & Attention Tracing & Grad-SAM \\
  \midrule
  ML-1M ($L{=}200$)             & 6.9864  & 0.8791  & 0.1459 & 0.4405  \\
  ML-10M ($L{=}500$)            & 26.9987 & 3.1202  & 0.5183 & 1.5628  \\
  KuaiRand-1K ($L{=}500$)       & 38.7506 & 4.2417  & 0.5272 & 36.8118 \\
  KuaiRand-1K-Long ($L{=}1000$) & 97.0574 & 10.3945 & 1.5665 & 38.8862 \\
  \bottomrule
  \end{tabular}
  \end{table}

\section{Additional Results}
\begin{figure}[H]
      \centering
      \includegraphics[width=\textwidth]{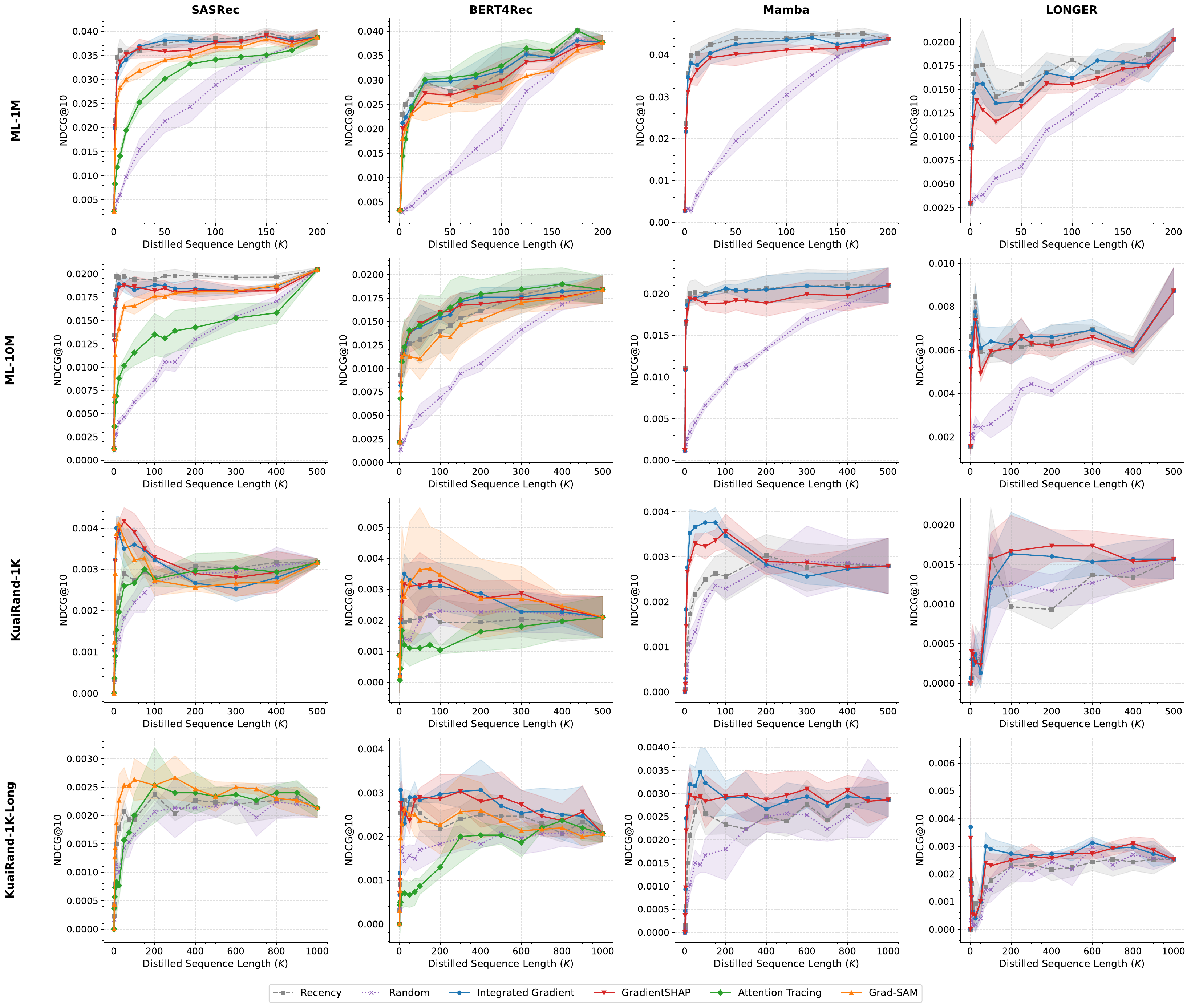}
      \caption{\textbf{ NDCG@10 vs.\ distilled sequence length $K$} for
  SASRec, BERT4Rec, Mamba, and LONGER (columns, left to right; standard
  attention only where applicable) evaluated on ML-1M, ML-10M, KuaiRand-1K, and
  KuaiRand-1K-Long (rows, top to bottom). Each curve shows
  $\text{NDCG@10}(K)/\text{NDCG@10}(K{=}L)$, i.e.\ performance at truncation
  length $K$ by the full-sequence ($K{=}L$) score, averaged across
  seeds; shaded regions denote $\pm 1$ standard deviation across seeds. Mamba
  and LONGER have no explicit attention matrix, so Attention Tracing and
  Attention~$\times$~Gradient are omitted for those columns.
      }
      \label{fig:app_arch_ndcg10}
  \end{figure}

\begin{figure}[H]
      \centering
      \includegraphics[width=\textwidth]{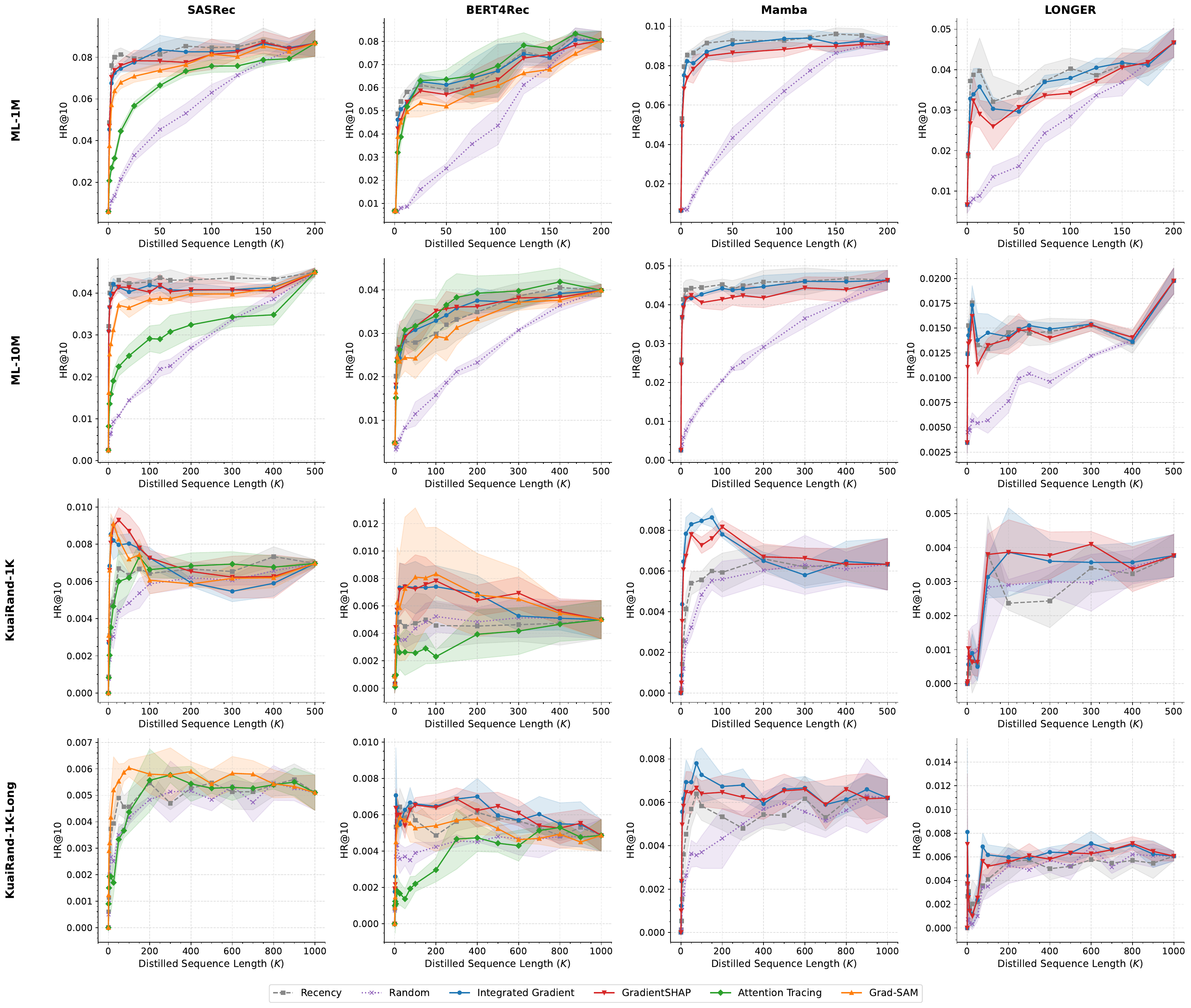}
      \caption{\textbf{ HR@10 vs.\ distilled sequence length $K$} for
  SASRec, BERT4Rec, Mamba, and LONGER (columns, left to right; standard
  attention only where applicable) evaluated on ML-1M, ML-10M, KuaiRand-1K, and
  KuaiRand-1K-Long (rows, top to bottom). Each curve shows
  $\text{HR@10}(K)/\text{HR@10}(K{=}L)$, i.e.\ performance at truncation
  length $K$  by the full-sequence ($K{=}L$) score, averaged across
  seeds; shaded regions denote $\pm 1$ standard deviation across seeds. Mamba
  and LONGER have no explicit attention matrix, so Attention Tracing and
  Attention~$\times$~Gradient are omitted for those columns.
      }
      \label{fig:app_arch_hr10}
  \end{figure}

\begin{figure}[H]
      \centering
      \includegraphics[width=\textwidth]{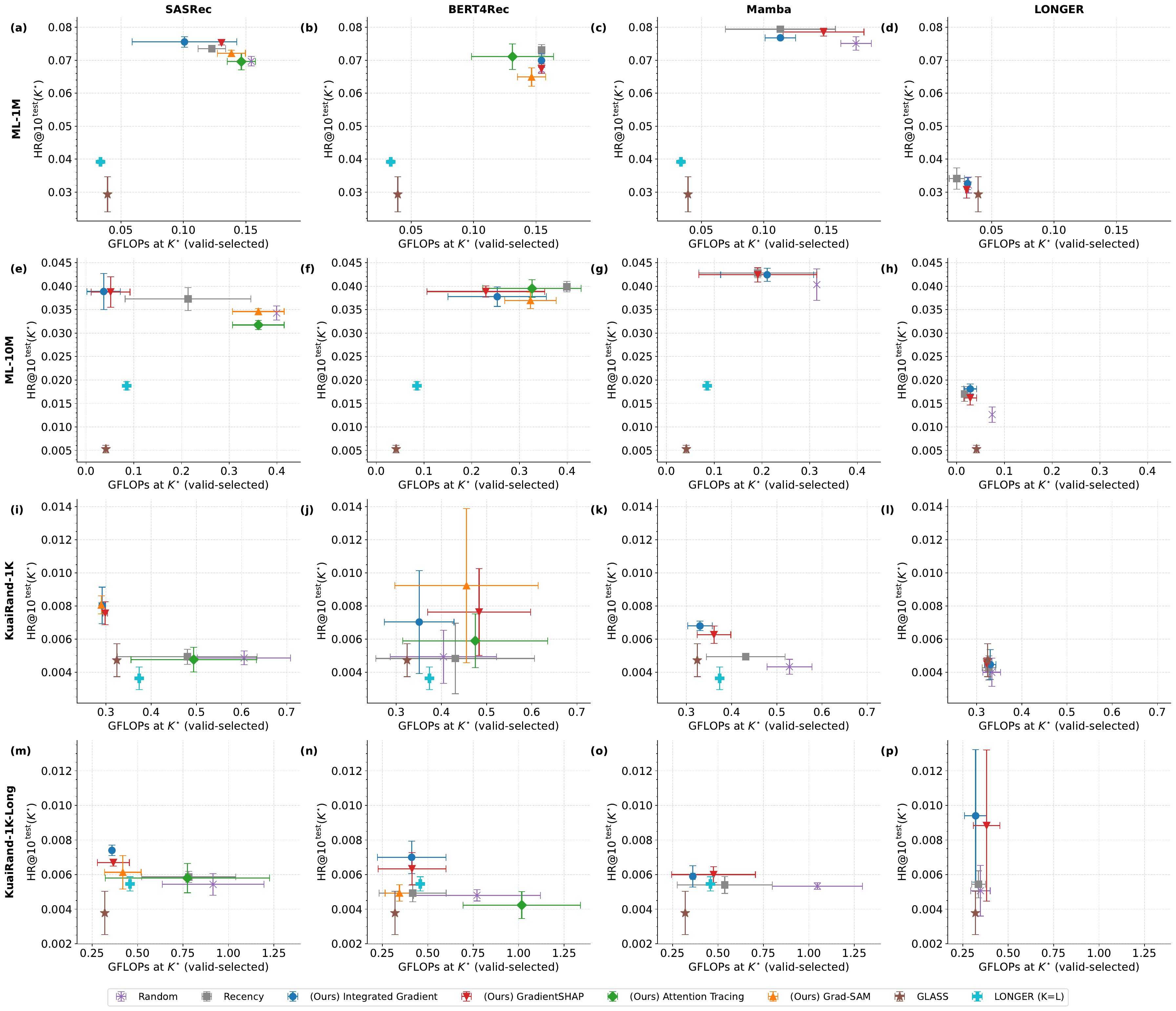}
      \vspace{-6mm}
      \caption{\textbf{Peak HR@10 vs.\  GFLOPs} for SASRec  (a, e, i, m), BERT4Rec (b, f, j, n), Mamba (c, g, k, o), and LONGER (d, h, l,
  p), each with standard attention, evaluated on MovieLens-1M (a--d),  ML-10M (e--h), KuaiRand-1K (i--l), and KuaiRand-1K-Long (m--p). Points closer to the \textbf{upper-left corner} (higher NDCG@10 at lower  GFLOPs) indicate a \textbf{better performance-efficiency trade-off}.}
      \label{fig:pareto_sasrec_hr}
  \end{figure}

\begin{figure}[H]
      \centering
      \includegraphics[width=\textwidth]{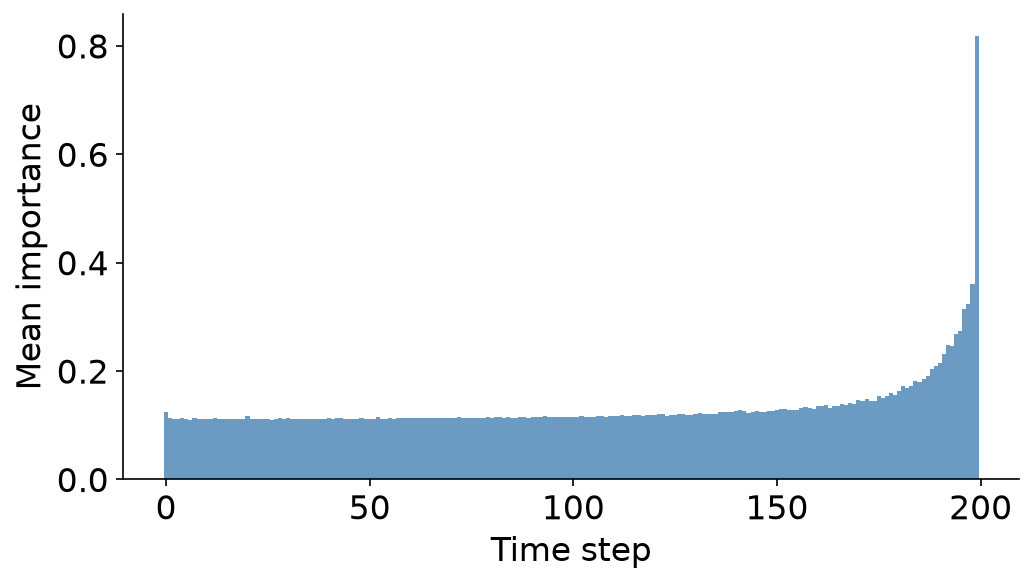}
      \caption{Importance score as function of temporal position, calculated from Grad-SAM method on ML-1M dataset. Higher the temporal position, more recent the point. The plot shows high recency bias.}
      \label{fig:recnecy_bias}
  \end{figure}

\begin{figure}[H]
\centering
\includegraphics[width=0.96\textwidth]{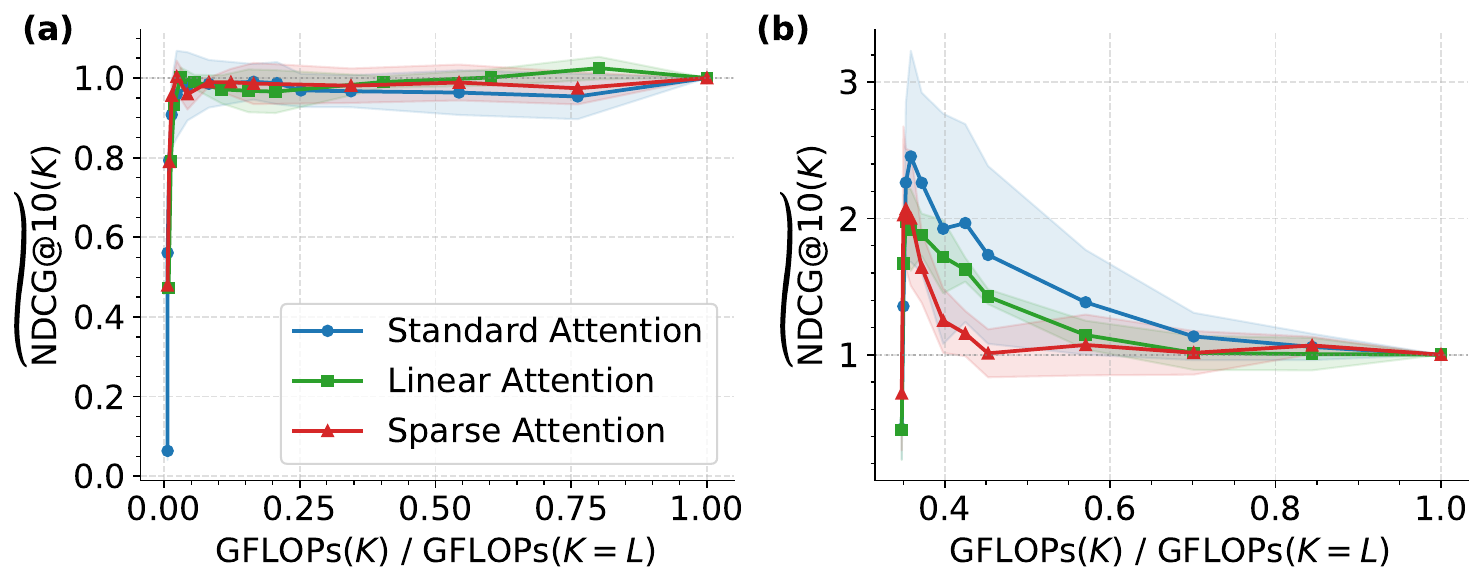}
\vspace{-3mm}
\caption{ Normalized $\widetilde{\mathrm{NDCG}@10}(K)$ vs.\ normalized GFLOPs for SASRec with Integrated Gradients on ML-10M (a) and KuaiRand-1K (b) across attention variants. }
\label{fig:sasrec_ig_attn_variants}
\end{figure} 

\subsection{Statistical Analysis} \label{statstical_analysis}

\begin{figure}[H]
      \centering
      \includegraphics[width=\textwidth]{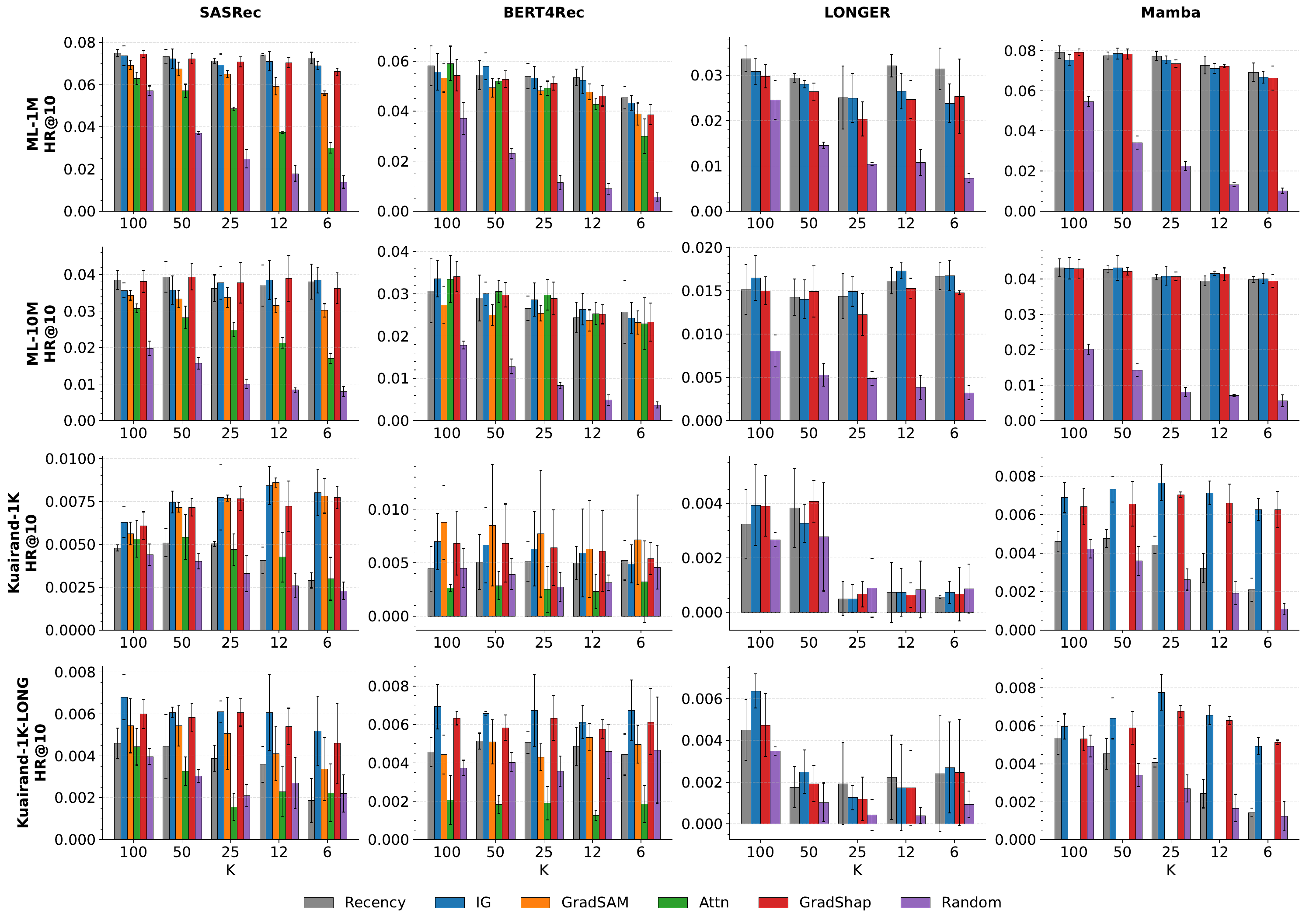}
       \caption{\textbf{HR@10 under sequence distillation across datasets and architectures}. Test HR@10 of each distillation method as a function of the distilled sequence length $K \in \{100, 50, 25, 12, 6\}$.}
      \label{fig:bars_hr}
  \end{figure}

\begin{figure}[H]
      \centering
      \includegraphics[width=\textwidth]{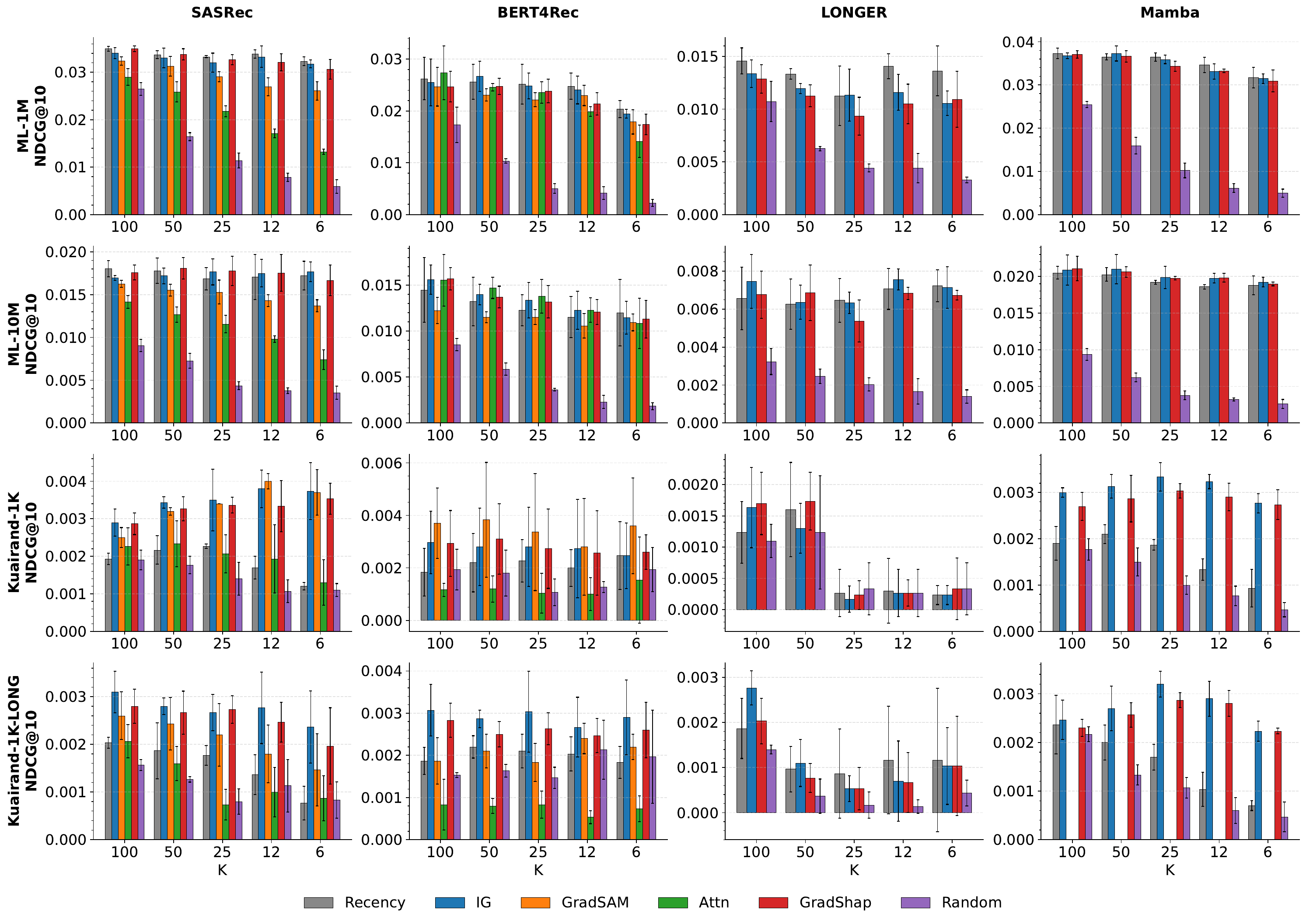}
      \caption{\textbf{NDCG@10 under sequence distillation across datasets and architectures}. Test NDCG@10 of each distillation method as a function of the distilled sequence length $K \in \{100, 50, 25, 12, 6\}$.}
      \label{fig:bars_ndcg}
  \end{figure}

Unlike the peak-$K^\star$ analysis, which contrasts unconstrained empirical optima, this evaluation compares methods under \emph{strictly matched computational budgets}: every method is constrained to identical sequence lengths $K$, isolating selection quality from operating point bias. The comparison at different $K$ values is shown in Figure \ref{fig:bars_hr} and Figure \ref{fig:bars_ndcg}. Across all evaluation configurations ($\text{seed} \times K \times \text{architecture}$), we compute the mean Friedman rank for each distillation method (Tables~\ref{tab:friedman_rank} and~\ref{rank_hr}; lower is better). 

A Friedman test strongly rejects the null hypothesis of equivalent performance across both metrics ($p < 10^{-60}$). Integrated Gradients (IG) achieve the strongest overall rankings, significantly outperforming heuristic selection well beyond the Nemenyi critical difference threshold ($\mathrm{CD}_{0.05} = 0.69$). 

Crucially, the relative ordering between IG and Recency reveals clear domain-specific dynamics:
\begin{itemize}
    \item \textbf{Compact, Recency-Dominated Sequences (ML-1M):} Recency acts as an competitive inductive bias, outranking IG ($\mathrm{NDCG@10}$ rank $1.68$ vs.\ $2.08$), reflecting the strong temporal locality inherent to movie-watching behavior.
    \item \textbf{Transition to Larger Catalogs (ML-10M):} This relationship reverses as catalog diversity scales; IG overtakes Recency with an $\mathrm{NDCG@10}$ rank of $2.18$ versus $2.73$.
    \item \textbf{High-Noise, Long Sequences (KuaiRand):} The divergence becomes pronounced on KuaiRand-1K and KuaiRand-1K-LONG, where Recency degrades to mid-tier performance (ranks $3.78\text{--}4.07$), while IG consistently dominates top ranks ($1.35\text{--}2.52$).
\end{itemize}
Across all evaluated budgets, IG demonstrates superior stability, securing the top overall rank on both metrics while never dropping below second on any individual benchmark.

\begin{table}[!htbp]
    \centering
    \small
    \caption{\textbf{Mean Friedman rank by $\mathrm{NDCG@10}$ (lower is better)}. For each method, the mean rank is calculated across 3 seeds, 5 different $K$ values, and 4 different architectures.}
    \label{tab:friedman_rank}
    \begin{tabular}{lrrrrr}
    \toprule
     & ML-1M & ML-10M & KuaiRand-1K & KuaiRand-1K-LONG & Overall \\
    \midrule
    Integrated Gradient & \underline{2.08} & \textbf{2.18} & \underline{2.35} & \textbf{1.40} & \textbf{2.00} \\
    GradientSHAP        & 3.02             & \underline{2.38} & 2.58             & \underline{2.00} & \underline{2.50} \\
    Recency             & \textbf{1.68}    & 2.73             & 3.97             & 4.07             & 3.11 \\
    Grad-SAM            & 4.00             & 4.13             & \textbf{2.07}    & 3.23             & 3.36 \\
    Attention Tracing   & 4.22             & 3.57             & 4.83             & 5.43             & 4.51 \\
    Random              & 6.00             & 6.00             & 5.20             & 4.87             & 5.52 \\
    \bottomrule
    \end{tabular}
\end{table}

\begin{table}[!htbp]
    \centering
    \small
    \caption{\textbf{Mean Friedman rank by $\mathrm{HR@10}$ (lower is better)}. For each method, the mean rank is calculated across 3 seeds, 5 different $K$ values, and 4 different architectures.}
    \label{rank_hr}
    \begin{tabular}{lrrrrr}
    \toprule
     & ML-1M & ML-10M & KuaiRand-1K & KuaiRand-1K-LONG & Overall \\
    \midrule
    Integrated Gradient & \underline{2.15} & \textbf{2.27}    & \underline{2.52} & \textbf{1.35}    & \textbf{2.07} \\
    GradientSHAP        & 2.97             & 2.60             & 2.60             & \underline{1.97} & \underline{2.53} \\
    Recency             & \textbf{1.68}    & \underline{2.53} & 3.78             & 4.02             & 3.00 \\
    Grad-SAM            & 3.87             & 3.95             & \textbf{2.05}    & 3.22             & 3.27 \\
    Attention Tracing   & 4.33             & 3.65             & 4.83             & 5.62             & 4.61 \\
    Random              & 6.00             & 6.00             & 5.22             & 4.83             & 5.51 \\
    \bottomrule
    \end{tabular}
\end{table}

\begin{table}[!htbp]
    \centering
    \footnotesize
    \renewcommand{\arraystretch}{0.8} 
    \setlength{\tabcolsep}{4pt}          
    \caption{$\mathrm{AUC}_{\mathrm{HR}@10}$ (mean $\pm$ std across seeds) across different model architectures on $\mathcal{D}_{\mathrm{valid}}$.}
    \label{tab:hr_combined}
    \begin{adjustbox}{max width=\textwidth}
    \begin{tabular}{llcccc}
    \toprule
    Method & Submethod & ML-1M & ML-10M & KuaiRand-1K & KuaiRand-1K-Long \\
    \midrule
    \multicolumn{6}{l}{\textbf{\textit{SASRec}}} \\
    Random               & -- & $0.697 \pm 0.024$ & $0.714 \pm 0.074$ & $0.900 \pm 0.068$ & $0.90 \pm 0.12$ \\
    Recency              & -- & $\underline{0.983 \pm 0.037}$ & $\mathbf{1.000 \pm 0.049}$ & $0.961 \pm 0.011$ & $1.000 \pm 0.098$ \\
    \multirow{4}{*}{\shortstack[l]{Ours\\(XAI-Driven)}} & \cellcolor{lightmaroon}Integrated Gradient & \cellcolor{lightmaroon}$\underline{0.983 \pm 0.063}$ & \cellcolor{lightmaroon}$0.972 \pm 0.044$ & \cellcolor{lightmaroon}$\mathbf{1.141 \pm 0.061}$ & \cellcolor{lightmaroon}$1.04 \pm 0.10$ \\
     & \cellcolor{lightmaroon}GradientSHAP      & \cellcolor{lightmaroon}$\mathbf{0.985 \pm 0.044}$ & \cellcolor{lightmaroon}$\underline{0.988 \pm 0.023}$ & \cellcolor{lightmaroon}$\underline{1.093 \pm 0.065}$ & \cellcolor{lightmaroon}$\underline{1.06 \pm 0.14}$ \\
     & \cellcolor{lightmaroon}Attention Tracing & \cellcolor{lightmaroon}$0.807 \pm 0.044$ & \cellcolor{lightmaroon}$0.836 \pm 0.050$ & \cellcolor{lightmaroon}$0.951 \pm 0.029$ & \cellcolor{lightmaroon}$0.99 \pm 0.12$ \\
     & \cellcolor{lightmaroon}Grad-SAM          & \cellcolor{lightmaroon}$0.921 \pm 0.040$ & \cellcolor{lightmaroon}$0.926 \pm 0.078$ & \cellcolor{lightmaroon}$1.043 \pm 0.063$ & \cellcolor{lightmaroon}$\mathbf{1.097 \pm 0.094}$ \\
    \addlinespace[2pt]
    \multicolumn{6}{l}{\textbf{\textit{BERT4Rec}}} \\
    Random               & -- & $0.587 \pm 0.043$ & $0.728 \pm 0.033$ & $1.025 \pm 0.057$ & $1.15 \pm 0.14$ \\
    Recency              & -- & $\mathbf{0.895 \pm 0.064}$ & $0.904 \pm 0.040$ & $1.011 \pm 0.097$ & $1.14 \pm 0.14$ \\
    \multirow{4}{*}{\shortstack[l]{Ours\\(XAI-Driven)}} & \cellcolor{lightmaroon}Integrated Gradient & \cellcolor{lightmaroon}$\underline{0.877 \pm 0.060}$ & \cellcolor{lightmaroon}$\underline{0.915 \pm 0.007}$ & \cellcolor{lightmaroon}$1.32 \pm 0.38$ & \cellcolor{lightmaroon}$\mathbf{1.41 \pm 0.34}$ \\
     & \cellcolor{lightmaroon}GradientSHAP      & \cellcolor{lightmaroon}$0.837 \pm 0.057$ & \cellcolor{lightmaroon}$0.912 \pm 0.029$ & \cellcolor{lightmaroon}$\underline{1.40 \pm 0.35}$ & \cellcolor{lightmaroon}$\underline{1.32 \pm 0.22}$ \\
     & \cellcolor{lightmaroon}Attention Tracing & \cellcolor{lightmaroon}$0.870 \pm 0.042$ & \cellcolor{lightmaroon}$\mathbf{0.941 \pm 0.027}$ & \cellcolor{lightmaroon}$0.89 \pm 0.12$ & \cellcolor{lightmaroon}$0.93 \pm 0.18$ \\
     & \cellcolor{lightmaroon}Grad-SAM          & \cellcolor{lightmaroon}$0.818 \pm 0.037$ & \cellcolor{lightmaroon}$0.836 \pm 0.011$ & \cellcolor{lightmaroon}$\mathbf{1.69 \pm 0.68}$ & \cellcolor{lightmaroon}$1.23 \pm 0.31$ \\
    \addlinespace[2pt]
    \multicolumn{6}{l}{\textbf{\textit{Mamba}}} \\
    Random               & -- & $0.650 \pm 0.019$ & $0.682 \pm 0.031$ & $0.90 \pm 0.11$ & $0.89 \pm 0.12$ \\
    Recency              & -- & $\mathbf{0.976 \pm 0.030}$ & $\underline{0.983 \pm 0.004}$ & $0.99 \pm 0.11$ & $0.955 \pm 0.099$ \\
    \multirow{2}{*}{\shortstack[l]{Ours\\(XAI-Driven)}} & \cellcolor{lightmaroon}Integrated Gradient & \cellcolor{lightmaroon}$0.948 \pm 0.027$ & \cellcolor{lightmaroon}$\mathbf{0.983 \pm 0.031}$ & \cellcolor{lightmaroon}$\mathbf{1.30 \pm 0.19}$ & \cellcolor{lightmaroon}$\underline{1.04 \pm 0.14}$ \\
     & \cellcolor{lightmaroon}GradientSHAP      & \cellcolor{lightmaroon}$\underline{0.962 \pm 0.033}$ & \cellcolor{lightmaroon}$0.971 \pm 0.012$ & \cellcolor{lightmaroon}$\underline{1.20 \pm 0.19}$ & \cellcolor{lightmaroon}$\mathbf{1.05 \pm 0.12}$ \\
    \addlinespace[2pt]
    \multicolumn{6}{l}{\textbf{\textit{LONGER}}} \\
    Random               & -- & $0.567 \pm 0.015$ & $0.579 \pm 0.051$ & $0.776 \pm 0.063$ & $0.795 \pm 0.076$ \\
    Recency              & -- & $\mathbf{0.830 \pm 0.006}$ & $\underline{0.815 \pm 0.082}$ & $0.83 \pm 0.19$ & $0.854 \pm 0.086$ \\
    \multirow{2}{*}{\shortstack[l]{Ours\\(XAI-Driven)}} & \cellcolor{lightmaroon}Integrated Gradient & \cellcolor{lightmaroon}$\underline{0.763 \pm 0.018}$ & \cellcolor{lightmaroon}$\mathbf{0.819 \pm 0.048}$ & \cellcolor{lightmaroon}$\underline{1.00 \pm 0.14}$ & \cellcolor{lightmaroon}$\mathbf{1.09 \pm 0.16}$ \\
     & \cellcolor{lightmaroon}GradientSHAP      & \cellcolor{lightmaroon}$0.728 \pm 0.028$ & \cellcolor{lightmaroon}$0.773 \pm 0.043$ & \cellcolor{lightmaroon}$\mathbf{1.03 \pm 0.11}$ & \cellcolor{lightmaroon}$\underline{1.018 \pm 0.086}$ \\
    \bottomrule
    \end{tabular}
    \end{adjustbox}
\end{table}

\begin{figure}[H]
      \centering
      \includegraphics[width=\textwidth]{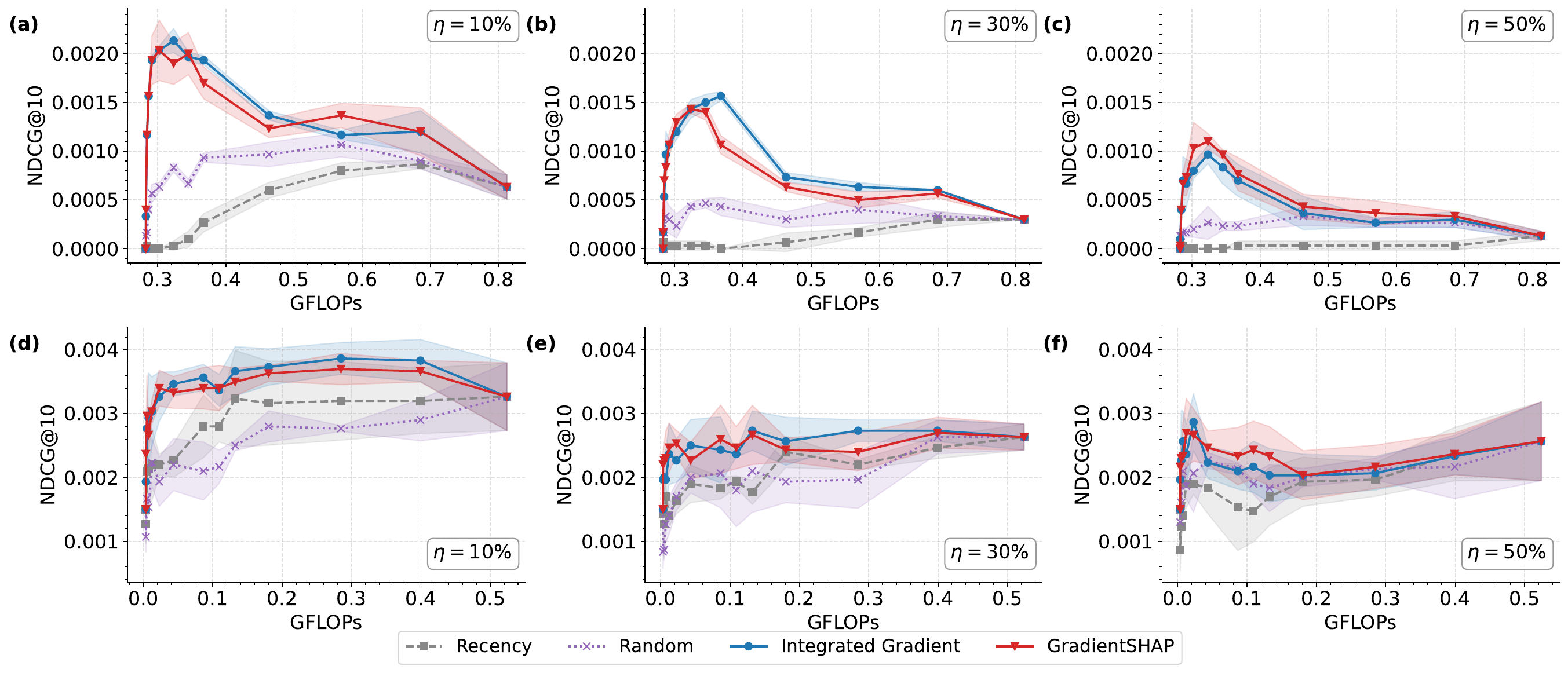}
      \caption{NDCG@10 as a function of GFLOPs for SASRec on KuaiRand-1K (a-c) and ML-1M (d-f) under \textbf{recent} item noise injection at three noise levels ($\eta \in \{10\%, 30\%, 50\%\}$). The \textbf{rightmost point} in each graph represents the performance with \textbf{no distillation}.}
      \label{fig:recency_noise_ablation}
  \end{figure}
  


\newpage
\section*{NeurIPS Paper Checklist}

The checklist is designed to encourage best practices for responsible machine learning research, addressing issues of reproducibility, transparency, research ethics, and societal impact. Do not remove the checklist: {\bf The papers not including the checklist will be desk rejected.} The checklist should follow the references and follow the (optional) supplemental material.  The checklist does NOT count towards the page
limit. 

Please read the checklist guidelines carefully for information on how to answer these questions. For each question in the checklist:
\begin{itemize}
    \item You should answer \answerYes{}, \answerNo{}, or \answerNA{}.
    \item \answerNA{} means either that the question is Not Applicable for that particular paper or the relevant information is Not Available.
    \item Please provide a short (1--2 sentence) justification right after your answer (even for \answerNA). 
\end{itemize}

{\bf The checklist answers are an integral part of your paper submission.} They are visible to the reviewers, area chairs, senior area chairs, and ethics reviewers. You will also be asked to include it (after eventual revisions) with the final version of your paper, and its final version will be published with the paper.

The reviewers of your paper will be asked to use the checklist as one of the factors in their evaluation. While \answerYes{} is generally preferable to \answerNo{}, it is perfectly acceptable to answer \answerNo{} provided a proper justification is given (e.g., error bars are not reported because it would be too computationally expensive'' or ``we were unable to find the license for the dataset we used''). In general, answering \answerNo{} or \answerNA{} is not grounds for rejection. While the questions are phrased in a binary way, we acknowledge that the true answer is often more nuanced, so please just use your best judgment and write a justification to elaborate. All supporting evidence can appear either in the main paper or the supplemental material, provided in appendix. If you answer \answerYes{} to a question, in the justification please point to the section(s) where related material for the question can be found.

IMPORTANT, please:
\begin{itemize}
    \item {\bf Delete this instruction block, but keep the section heading ``NeurIPS Paper Checklist"},
    \item  {\bf Keep the checklist subsection headings, questions/answers and guidelines below.}
    \item {\bf Do not modify the questions and only use the provided macros for your answers}.
\end{itemize}


\begin{enumerate}

\item {\bf Claims}
    \item[] Question: Do the main claims made in the abstract and introduction accurately reflect the paper's contributions and scope?
    \item[] Answer: \answerYes{} 
    \item[] Justification: Results given in Section 4 and Appendix accurately reflect our claims.

\item {\bf Limitations}
    \item[] Question: Does the paper discuss the limitations of the work performed by the authors?
    \item[] Answer: \answerYes{} 
    \item[] Justification: In conclusion we have discussed the future direction. Also in the Section 3.1, we clearly explain the scope of our work.

\item {\bf Theory assumptions and proofs}
    \item[] Question: For each theoretical result, does the paper provide the full set of assumptions and a complete (and correct) proof?
    \item[] Answer: \answerNA{} 
    \item[] Justification: \answerNA{}

    \item {\bf Experimental result reproducibility}
    \item[] Question: Does the paper fully disclose all the information needed to reproduce the main experimental results of the paper to the extent that it affects the main claims and/or conclusions of the paper (regardless of whether the code and data are provided or not)?
    \item[] Answer: \answerYes{} 
    \item[] Justification: Details for reproducibility given in Section 3 and  Appendix A. 

\item {\bf Open access to data and code}
    \item[] Question: Does the paper provide open access to the data and code, with sufficient instructions to faithfully reproduce the main experimental results, as described in supplemental material?
    \item[] Answer: \answerYes{} 
    \item[] Justification: Authors will open-source the code upon acceptance.

\item {\bf Experimental setting/details}
    \item[] Question: Does the paper specify all the training and test details (e.g., data splits, hyperparameters, how they were chosen, type of optimizer) necessary to understand the results?
    \item[] Answer: \answerYes{} 
    \item[] Justification: Given in Section 3 and Appendix A.

\item {\bf Experiment statistical significance}
    \item[] Question: Does the paper report error bars suitably and correctly defined or other appropriate information about the statistical significance of the experiments?
    \item[] Answer: \answerYes{} 
    \item[] Justification: Results presented in section 3 and Appendix contain mean and standard deviation across 3 seeds.

\item {\bf Experiments compute resources}
    \item[] Question: For each experiment, does the paper provide sufficient information on the computer resources (type of compute workers, memory, time of execution) needed to reproduce the experiments?
    \item[] Answer: \answerYes{} 
    \item[] Justification: Provided in Appendix A.
    
\item {\bf Code of ethics}
    \item[] Question: Does the research conducted in the paper conform, in every respect, with the NeurIPS Code of Ethics \url{https://neurips.cc/public/EthicsGuidelines}?
    \item[] Answer: \answerYes{} 
    \item[] Justification: Authors have followed all the ethics guideline while carrying out this study.

\item {\bf Broader impacts}
    \item[] Question: Does the paper discuss both potential positive societal impacts and negative societal impacts of the work performed?
    \item[] Answer: \answerNA{} 
    \item[] Justification: \answerNA{} 
    
\item {\bf Safeguards}
    \item[] Question: Does the paper describe safeguards that have been put in place for responsible release of data or models that have a high risk for misuse (e.g., pre-trained language models, image generators, or scraped datasets)?
    \item[] Answer: \answerNA{}  
    \item[] Justification: \answerNA{} 

\item {\bf Licenses for existing assets}
    \item[] Question: Are the creators or original owners of assets (e.g., code, data, models), used in the paper, properly credited and are the license and terms of use explicitly mentioned and properly respected?
    \item[] Answer: \answerYes{} 
    \item[] Justification: All datasets methods used are cited in Section 3 and 4. 

\item {\bf New assets}
    \item[] Question: Are new assets introduced in the paper well documented and is the documentation provided alongside the assets?
    \item[] Answer: \answerNA{} 
    \item[] Justification: \answerNA{} 

\item {\bf Crowdsourcing and research with human subjects}
    \item[] Question: For crowdsourcing experiments and research with human subjects, does the paper include the full text of instructions given to participants and screenshots, if applicable, as well as details about compensation (if any)? 
    \item[] Answer: \answerNA{}  
    \item[] Justification: \answerNA{} 

\item {\bf Institutional review board (IRB) approvals or equivalent for research with human subjects}
    \item[] Question: Does the paper describe potential risks incurred by study participants, whether such risks were disclosed to the subjects, and whether Institutional Review Board (IRB) approvals (or an equivalent approval/review based on the requirements of your country or institution) were obtained?
    \item[] Answer: \answerNA{}  
    \item[] Justification: \answerNA{} 

\item {\bf Declaration of LLM usage}
    \item[] Question: Does the paper describe the usage of LLMs if it is an important, original, or non-standard component of the core methods in this research? Note that if the LLM is used only for writing, editing, or formatting purposes and does \emph{not} impact the core methodology, scientific rigor, or originality of the research, declaration is not required.
    \item[] Answer: \answerYes{} 
    \item[] Justification: Authors have used LLM for editing and plotting purposes.

\end{enumerate}

\end{document}